\documentclass[lettersize,journal]{IEEEtran}
\usepackage{amsmath,amsfonts}
\usepackage{algorithmic}
\usepackage{array}
\usepackage[caption=false,font=normalsize,labelfont=sf,textfont=sf]{subfig}
\usepackage{textcomp}
\usepackage{stfloats}
\usepackage{url}
\usepackage{verbatim}
\usepackage{graphicx}
\def\BibTeX{{\rm B\kern-.05em{\sc i\kern-.025em b}\kern-.08em
    T\kern-.1667em\lower.7ex\hbox{E}\kern-.125emX}}
\usepackage{balance}

\usepackage{epsfig}
\RequirePackage{booktabs}

\begin{document}
	
\title{Full-Reference Calibration-Free \\ Image Quality Assessment}

\author{\thanks{Copyright \copyright 2010 IEEE. Personal use of this material is permitted. However, permission to use this material for any other purposes must be obtained from the IEEE by sending a request to pubs-permissions@ieee.org.} Elio D. Di Claudio\thanks{Elio D. Di Claudio, Dept. of Information Engineering, Electronics and Telecommunications (DIET), University of Rome ``La Sapienza'', Via Eudossiana 18, I-00184 Rome, Italy. Ph.: +39-06-44585490, Fax: +39-06-44585632, e-mail elio.diclaudio@uniroma1.it.}, Paolo Giannitrapani\thanks{Contact author: Paolo Giannitrapani, Dept. of Information Engineering, Electronics and Telecommunications (DIET), University of Rome ``Sapienza'', Via Eudossiana 18, I-00184 Rome, Italy. Ph.: +39-06-44585490, Fax: +39-06-44585632, e-mail paolo.giannitrapani@uniroma1.it.} and Giovanni Jacovitti\thanks{Giovanni Jacovitti, formerly with Dept. of Information Engineering, Electronics and Telecommunications (DIET), University of Rome ``La Sapienza'', Via Eudossiana 18, I-00184 Rome, Italy. Ph.: +39-06-44585838, Fax: +39-06-44585632, e-mail gianni.iacovitti@gmail.com.}\thanks{EDICS No.: SMR-HPM.}}

\markboth{IEEE Transactions on Image Processing,~Vol.~xx, No.~xx, xx~xxxx}%
{Full-Reference Calibration-Free Image Quality Assessment}

\maketitle

\begin{abstract}
One major problem of objective Image Quality Assessment (IQA) methods is the lack of linearity of their quality estimates with respect to scores expressed by human subjects. For this reason, usually IQA metrics undergo a calibration process based on subjective quality examples. However, example-based training makes generalization problematic, hampering result comparison across different applications and operative conditions. In this paper, new Full Reference (FR) techniques, providing estimates linearly correlated with human scores without using calibration are introduced. To reach this objective, these techniques are deeply rooted on principles and theoretical constraints. Restricting the interest on the IQA of the set of natural images, it is first recognized that application of estimation theory and psycho physical principles to images degraded by Gaussian blur leads to a so-called canonical IQA method, whose estimates are not only highly linearly correlated to subjective scores, but are also straightforwardly related to the Viewing Distance (VD). Then, it is shown that mainstream IQA methods can be reconducted to the canonical method applying a preliminary  metric conversion based on a unique specimen image. The application of this scheme is then extended to a significant class of degraded images other than Gaussian blur, including noisy and compressed images. The resulting calibration-free FR IQA methods are suited for applications where comparability and interoperability across different imaging systems and on different VDs is a major requirement. A comparison of their statistical performance with respect to some conventional calibration prone methods is finally provided.
\end{abstract}


\section{Introductive notes}
\IEEEPARstart{I}{mage} Quality Assessment (IQA) is based on \emph{subjective} as well as \emph{objective} methods.

The \emph{subjective quality} of an image is defined as the average quality score assigned to it by a reference class of human subjects, usually expressed in the MOS (Mean Opinion Score) scale. Likewise, the \emph{subjective quality loss} with respect to a pristine version of the same image considered of perfect quality (also referred  to as reference or original image) is expressed using  DMOS (Difference of Mean Opinion Score) units.

The \emph{objective quality} consists instead of the algorithmic prediction of the subjective quality based on measurable image features, expressed as well in MOS/DMOS units.

In the recent decades, many objective quality assessment methods have been proposed. Methods requiring a complete representation of the reference image are referred to as Full Reference (FR) IQA methods, while methods using incomplete representations are referred to as Reduced Reference (RR). Methods based only on the knowledge of degraded images are finally referred to as No Reference (NR) methods. The present paper concerns FR methods.

The typical scheme of objective FR IQA methods includes first a local comparative analysis of corresponding details of the  pristine and of the degraded images, followed by a pooling stage over the whole image. The result is referred to as IQA \emph{metric} (sometimes termed “objective quality” for short).

Formulation of existing metrics was inspired by different criteria. Some metrics measure the similarity among image representations (reproduction fidelity) \cite{MANNOS74,WANG04,FSIMPAGE13,XUE14}, possibly accounting for constraints suggested by more or less sophisticated models of the Human Visual System (HVS). Other classical metrics measure a loss of a somehow defined \emph{visual information} caused by the image degradation, with explicit reference to a cognitive interpretation of the HVS role \cite{SHEIKH06,VQEG00}. Metrics have  mostly the form of scalar metrics, even if some vector metrics have been adopted to distinguish among different types of image degradation \cite{DICLAUDIO18}.

The final step is the application of a scoring function whose goal is to convert the IQA metric values into the subjective MOS/DMOS scales (for short, the result is sometimes referred to as “subjective quality”, subtending “estimates of”). The scoring function is typically formulated as a parametric function suggested by the aim to model threshold and saturation phenomena. Parameters are adjusted by non-linear regression using available empirical examples. Following the VQEG suggestions \cite{VQEG00,VQEG03}, the scoring function for a scalar metric consists of a logistic function whose typical form is
\begin{equation}
	\hat{m}(\zeta)=\beta_1\left[\frac{1}{2}-\frac{1}{1+e^{\beta_2\left(\zeta-\beta_3\right)}}\right]+\beta_4\zeta+\beta_5
	\label{eq:VQEG}
\end{equation}
where $\zeta$ is the metric value, $\hat{m}(\zeta)$ is the estimated DMOS value \cite{SHEIKH06B}, and $\beta_i$ are parameters usually adjusted minimizing the Euclidean distance of empirical DMOS values from the estimated ones.

Of course, the larger the number of parameters, the more accurate is the fitting of the specific training examples during calibration. At the same time the variability across different sets of examples grows with the number of parameters (\emph{overfitting}). Notice that in \cite{ITU16B} a S-shaped function regulated by three parameters only is advocated.

As a matter of fact, it is hard to find default values for calibration parameters in the available literature. While compensating for the variability of methodologies employed for subjective quality example collection, calibration prevents the fair comparison of quality measurement across different applications \cite{AKAIKE74}.

The calibration parameters critically depend on many factors, different protocols, different methods of determining the MOS/DMOS values from raw data, different experimental settings such as display contrast, room illumination, etc., besides local minima issues of the non-linear optimization process.

Most evidently, they depend on the Viewing Distance (VD) \cite{GU15}. In fact, the objective quality is calculated using the images reproduced on the display, while the subjective quality depends on the images perceived on the retina, whose scale is determined by the VD. Unfortunately, the parametric forms of scoring functions generally lack an explicit dependence of VD. In some IQA methods, namely the Multi-Scale Structural Similarity (MSSIM) \cite{WANG03} and the Feature Similarity Index (FSIM) \cite{FSIMPAGE13}, the dependence of their scoring function on the VD is attenuated by adjusting a mix of metrics calculated at different scales. Generalizing this approach, in \cite{GU15} a preprocessing stage is applied to different IQA methods. Images are rescaled to emulate different VDs, and the corresponding metrics are combined using the criterion of optimizing the overall statistical performance. The VD was also included, as a learnable parameter, in a CNN based NR IQA method \cite{BOSSE18}.

To circumvent the cited shortcomings, generalizability is pursued resorting to physical and psycho-physical theoretical constraints. The analysis conducted here stems from a theory about the annoyance of blur in vision, developed in \cite{DICLAUDIO21}. This theory is based on the overarching principle that the Human Vision System (HVS) is optimal for fine pattern localization, given some macro-structural constraints. The logical consequence of this principle is that the variance of fine localization of the HVS is inter-subjective and calculable as the inverse of the Positional Fisher Information (PFI) (see \cite{NERI04} for a general account).

Under this perspective, an essential abstract model of the HVS composed of the optical tract and of the neural tract (from the retina up to the visual cortex) is adopted. The last one is modeled by a single, complex valued, Virtual Receptive Field (VRF) accounting at the same time for pattern orientation selectivity and for 2D spatial frequency selectivity. The VRF allows straightforward calculation of the PFI of image details. 

A substantial simplification of the analysis comes from restricting the analysis to the class of the so-called \emph{natural images}, i.e., the images usually seen by subjects during their everyday experience directly or through imaging devices. In the present paper natural images are modeled as elements of a random set whose ensemble radial spectral energy distribution decays as $\frac{1}{\rho^2}$, where $\rho$ indicates the radial coordinate of the 2D Fourier transform of the images \cite{TORRALBA03}. Notice that other characterizations of natural images are used. See for instance the Natural Scene Statistics (NSS), adopted in \cite{SHEIKH06}.

At this point, it is assumed that the perception of the quality loss follows the psycho-physical Weber-Fechner law with respect to losses of pattern localizability, viewed as stimulus. It follows that an estimate of the subjective quality loss in the case of Gaussian blurred images is expressed in a closed, unexpectedly simple form \cite{DICLAUDIO21} defining a \emph{canonical rating function} (CRF) of a \emph{canonical metric} measuring the amount of  blur. The  CRF  is characterized by only two significant and operative parameters: the VD and a scoring gain depending on conventional quality scores assigned to “anchor” images.

Experimental verifications of such a \emph{canonical IQA method} for Gaussian blurred natural images conducted on independent IQA databases confirmed a considerable linearity versus empirical MOS/DMOS measurements.

In the second part of the paper the application of other conventional IQA methods to the quality estimation of Gaussian blurred, natural images is first introduced. A  key result is that the rating function of a classical IQA metric can be decomposed into the cascade of a metric conversion rule from specific metrics to the canonical metric, and the CRF.

The final part of the paper is devoted to the extension of these theoretical results to other types of image degradation, beyond the specific case of Gaussian blur. The extension is based on  the consideration that IQA methods are purposely designed to estimate the subjective image quality loss, irrespective of the degradation cause. Therefore, it is expected that different types of degradations will properly map onto the canonical metric provided they are perceptually equivalent, in terms of image quality, to a certain level of normalized Gaussian blur. The linearity of the overall scoring function follows as a logical consequence of the form of the CRF. Some mainstream IQA methods, namely the MSSIM, the FSIM, the VIF and the GSMD are specifically examined as possible candidates for rectification-free subjective quality prediction. They demonstrate the effectiveness of this approach for a technically significant class of image degradations, as shown by a comparative analysis of performance with empirically calibrated conventional methods, using independent databases.

The present paper is organized as follows. In Section II, an essential account of the employed HVS model is provided. In Section III, the canonical IQA method for Gaussian blurred images is illustrated. In Section IV, the performance of the canonical method for Gaussian blurred images is compared to that of other IQA methods. In Section V, the extension to other image degradation types is discussed. In Section VI, the performance of the calibration-free methods are compared to the popular competitors on different databases. Implementation issues are discussed in Section VII. Finally, some remarks are outlined  in Section VIII and conclusion is  drawn in Section IX.

\section{Abstract model of the human visual system}

\subsection{Optical tract}
\noindent The \emph{retina} in the foveal region is modeled as a distribution of light receptors whose position is individuated by the Cartesian coordinate pair $\mathbf{p}\equiv(x_1,x_2)$. The projection of an image on the retina generates a luminance component $I\left(\mathbf{p}\right)$ (the chrominance components are ignored here). It is assumed that $I\left(\mathbf{p}\right)$ is sampled by a grid of receptors whose average density is assumed as $60/degree$ of the angle of view. Since these receptors are not regularly placed on the retinal surface, invoking the generalized theory for non-baseband and nonuniform samples of \cite{LANDAU67}, this amounts to say that $I\left(\mathbf{p}\right)$ is correctly represented provided that its 2D bandwidth does not exceed $(-30,+30)$ $cycles/degree$, i.e., $(-1/2, +1/2)$ $cycles/arcmin$. This assumed retinal resolution corresponds to the so-called \emph{Snellen acuity}.

On the other hand, let us consider a display formed by a rectangular array of pixels characterized by its \emph{resolution} $R$ expressed in $pixel/mm$, i.e., spaced by $d=\frac{1}{R}$  $mm$ apart.

The image is projected onto the retina without loss of information and without redundancy if the spacing $d$ matches the said $60/degree$ density of receptors on the retina. This occurs at a VD $\delta_0$ such that:
\begin{equation}
	\delta_0\times\tan(1)=\frac{1}{R}=d
\end{equation}
where lengths are expressed in $mm$ and angles in $arcmins$. This distance is referred here to as \emph{nominal VD}. In the ITU recommendations \cite{ITU12} this distance is referred to as the \emph{Design Viewing Distance} (DVD), or \emph{Optimal Viewing Distance} (OVD).

Specifically, for a display of height $H$ (in $mm$) and characterized by $L$ rows of square pixel \emph{with reference to the center of the screen orthogonal to the line of sight}, we have (in $mm$):
\begin{equation}
	\delta_0=\frac{H}{L\tan(1)}=3438\times\frac{H}{L}
\end{equation}
For instance, for a 4K, 32 inch screen, the nominal VD in $mm$ is:
\begin{equation}
	\delta_0=3438\times\frac{440}{2160}=700
\end{equation}

At nominal VD the spatial Fourier spectra of the images projected onto the retina are limited into the range $(-1/2,+1/2)$ $cycles/arcmin$ or $(-30,+30)$ $cycles/degree$, according to the Nyquist criterion.

For a generic VD $\delta$, the spectrum of the retinal images lies within the band $\displaystyle(-\frac{1}{2}\frac{\delta}{\delta_0}, +\frac{1}{2}\frac{\delta}{\delta_0})$ $cycles/arcmin$ or $\displaystyle(-30\frac{\delta}{\delta_0}, +30\frac{\delta}{\delta_0})$ $cycles/degree$ both in horizontal and vertical directions. Thus, if the VD is less than the nominal one, the bandwidth of the sensed image shrinks. If the distance exceeds the nominal one, the bandwidth broadens.

\subsection{Neural tract}
\noindent The images captured by the retina are processed through successive stages of the neural tract connecting the output of the retina to the visual cortex, with \emph{receptive fields} distributed across the visual field \cite{HUBEL95}. In the present work, no attempt is made to model physical receptive fields. For the present purposes, it suffices to model the whole processing as the convolution of the sensed image $I\left(\mathbf{p}\right)$ with a single VRF \emph{model} \cite{DICLAUDIO21}, represented by its complex valued Point Spread Function (PSF) $h\left(\mathbf{p}\right)$:
\begin{IEEEeqnarray}{l}
	y\left(\mathbf{p}\right)=I\left(\mathbf{p}\right)\ast h\left(\mathbf{p}\right) \\
	h\left(\mathbf{p}\right)=Re{\left\{h\left(\mathbf{p}\right)\right\}}+jIm{\left\{h\left(\mathbf{p}\right)\right\}}
	\label{eqn:visualmap}
\end{IEEEeqnarray}

The output image $y\left(\mathbf{p}\right)$ will be referred to as \emph{Visual Map} (VM). The VRF is an abstract \emph{functional model} aimed to account at the same time for the \emph{orientation selective behavior} and the spatial \emph{frequency selective behavior} of the HVS according to the general principles of \cite{DAUGMAN83}.

In the polar coordinates $r=\sqrt{x_1^2+x_2^2}$ and $\displaystyle\varphi=tg^{-1}\frac{x_2}{x_1}$ it assumes the polar separable form:
\begin{equation}
	h(r,\varphi)=\frac{r}{2\pi s_G^2}e^{-\displaystyle\frac{r^2}{2{s_G}^2}}e^{j\varphi}
	\label{eqn:hmodel}
\end{equation}
where the parameter $s_G$ is the \emph{spread} of the VRF.

This VRF is a eigen-function with respect to 2D Fourier transformation \cite{DICLAUDIO10}, i.e., it maintains the same form. With reference to the polar coordinates in the frequency domain
\begin{equation}
	\rho=\sqrt{f_1^2+f_2^2}
\end{equation}
\begin{equation}
	\vartheta=tg^{-1}\frac{f_2}{f_1}
\end{equation}
where $\rho$ is the \emph{radial frequency}, the Fourier transform of the VRF is
\begin{equation}
	H(\rho,\vartheta)=j2\pi\rho e^{j\vartheta}e^{-S_G^2\rho^2}
\end{equation}

\begin{figure}[!ht]
	\centering
	\includegraphics[width=2.6in]{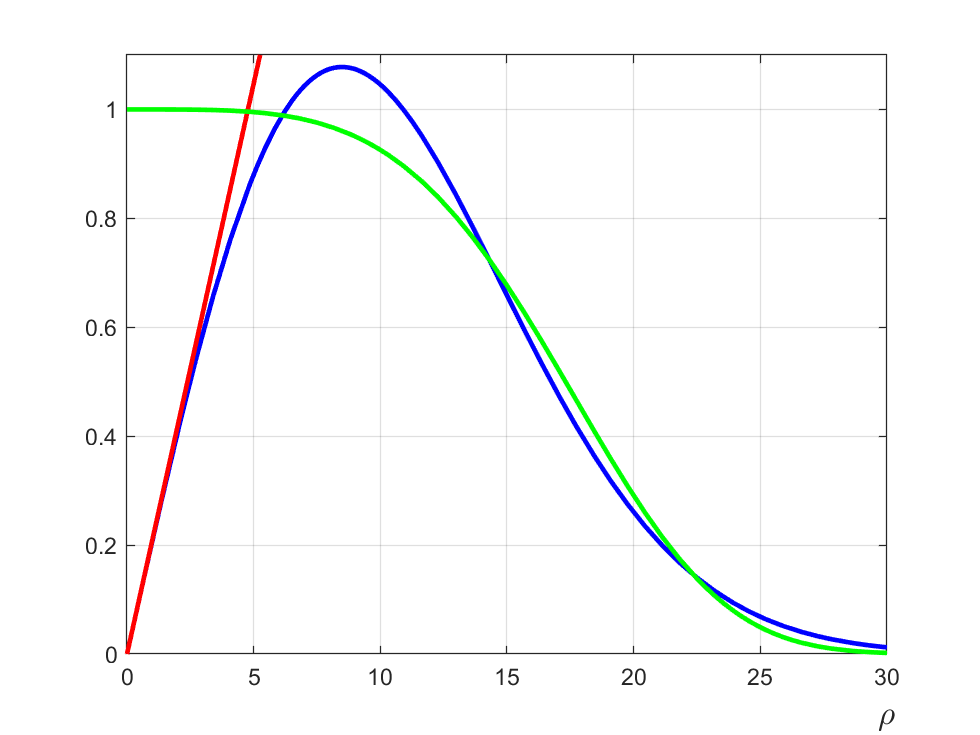}
	\caption{The radial frequency response magnitude of the VNTF for $s_G=2.5$ \emph{arcmins}. The maximum of the VNTF at about 8.5 \emph{cycles/degree}. It is the product of a linear component (red line) and a Gaussian-shaped component (green line).}
	\label{fig14}
\end{figure}

It is referred also to as Virtual Neural Transfer Function (VNTF). Its magnitude versus the radial frequency $\rho$ is a good model of the Contrast Sensitivity Function (CSF) of the HVS \cite{MANNOS74}, as shown in Fig.\ref{fig14} (blue line).

The maximum of the VNTF along the radial frequency is set at about $8.5$ \emph{cycles/degree}, according to the experimental data provided in the Fig.5 of \cite{CAMPBELL65}. This does correspond to a spread $s_G=2.5$ \emph{arcmin}, and a contrast loss of $\sim$40dB at the Nyquist frequency. The VNTF is naturally interpreted as the cascade of two functions. The first one is the magnitude of $j2\pi\rho e^{j\vartheta}$ (red line in Fig.\ref{fig14}), that represents the \emph{complex spatial gradient} operator \cite{REISERT08} defined as
\begin{equation}
	\nabla I\left(\mathbf{p}\right)\overset{\Delta}{=}\frac{\partial I\left(x_1,x_2\right)}{\partial x_1}+j\frac{\partial I\left(x_1,x_2\right)}{\partial x_2}
\end{equation}
This operator acts as an ideal edge extractor. The magnitude of its output $y\left(\mathbf{p}\right)$ indicates the local edge strengths, while its phase indicates the edge orientations.

The second factor (green line in Fig.\ref{fig14})
\begin{equation}
	G(\rho,\vartheta)=e^{-S_G^2\rho^2}
\end{equation}
represents a low-pass Gaussian-shaped filter acting as noise and aliasing suppressor. It is interpreted as the source of a \emph{neural blur}.

Now, let the image on the display $I_D\left(\mathbf{p}\right)$ be a blurred version of a pristine (non-degraded) image ${\widetilde{I}}_D\left(\mathbf{p}\right)$. It is modeled as the convolution with the blur kernel $ b\left(\mathbf{p}\right)$:
\begin{equation}
	I_D\left(\mathbf{p}\right)={\widetilde{I}}_D\left(\mathbf{p}\right)\ast b\left(\mathbf{p}\right)
\end{equation}
so that the blurred image on the retina is modeled as
\begin{equation}
	I(\mathbf{p})={\widetilde{I}}_D\left(\mathbf{p}\frac{\delta}{\delta_0}\right)\ast b\left(\mathbf{p}\frac{\delta}{\delta_0}\right)
\end{equation}
and finally the VM is modeled as
\begin{equation}
	y\left(\mathbf{p}\right)={\widetilde{I}}_D\left(\mathbf{p}\frac{\delta}{\delta_0}\right)\ast b\left(\mathbf{p}\frac{\delta}{\delta_0}\right)\ast h(\mathbf{p})
\end{equation}

\section{The canonical IQA method}
\label{sec:The canonical IQA method}
\noindent Accurate pattern localization is a primary goal of living beings. We assume \emph{as a principle} that the fine position of patterns in the observed images is determined by the HVS with the maximum allowable accuracy, given some macro-structural constraints. This leads one to maintain that such accuracy, measured by the inverse of the Fisher information, is \emph{inter-subjective}.

The PFI of a \emph{detail} of a pristine image, extracted by a window $w_\mathbf{p}(\mathbf{q})$, centered on $\mathbf{p}$ in presence of background white Gaussian noise with variance $\sigma_V^2$, is calculated in \cite{NERI04} as:
\begin{equation}
	\widetilde{\psi}\left(\mathbf{p}\right)=\frac{\widetilde{\lambda}\left(\mathbf{p}\right)}{\sigma_V^2}
\end{equation}
where
\begin{equation}
	\widetilde{\lambda}\left(\mathbf{p}\right)=\sum_{\mathbf{q}}{w_\mathbf{p}(\mathbf{q})^2\left|\widetilde{y}\left(\mathbf{p}-\mathbf{q}\right)\right|^2}
\end{equation}
is the energy of the detail. Likewise, for the same detail of a degraded version of the image we have
\begin{equation}
	\psi\left(\mathbf{p}\right)=\frac{\lambda\left(\mathbf{p}\right)}{\sigma_V^2}
\end{equation}
where
\begin{equation}
	\lambda\left(\mathbf{p}\right)=\sum_{\mathbf{q}}{w_\mathbf{p}(\mathbf{q})^2\left|y\left(\mathbf{p}-\mathbf{q}\right)\right|^2}
\end{equation}

Let us now calculate the PFI of the details in the frequency domain.  Here, a Gaussian blur applied to the observed image is described in polar frequency coordinates by the function
\begin{equation}
	B(\rho,\vartheta)=e^{-s_B^2\rho^2}
\end{equation}
where the parameter $s_B$ will be referred to as the \emph{spread} of the blur operator. Referring to the Fourier spectra of a generic detail, by the Parseval theorem we have\footnote{It is assumed that the shape of the detail window is so smooth that it does not influence significantly its spectrum.}:
\begin{IEEEeqnarray}{l}
	\psi\left(\mathbf{p}\right)=  \nonumber\\
	=\frac{1}{\sigma_V^2}\int_{0}^{2\pi}\int_{0}^{+\infty}{\left|D_\mathbf{p}\left(\rho\right)\right|^2\rho^2\left|G(\rho,\vartheta)\right|^2\left|B(\rho,\vartheta)\right|^2\rho d\rho d\vartheta} \nonumber\\
	\widetilde{\psi}\left(\mathbf{p}\right)=\frac{1}{\sigma_V^2}\int_{0}^{2\pi}\int_{0}^{+\infty}{\left|D_\mathbf{p}\left(\rho\right)\right|^2\rho^2\left|G(\rho,\vartheta)\right|^2\rho d\rho d\vartheta}
\end{IEEEeqnarray}

Let us now refer to \emph{natural images}. It is known that the expected value of the energy spectrum of details in the frequency domain is \cite{TORRALBA03}
\begin{equation}
	E\left\{\left|D_\mathbf{p}(\rho,\vartheta)\right|^2\right\}=f(\vartheta)\displaystyle\frac{1}{\rho^2}
\end{equation}

Therefore, for the random set of natural images, the expected value of the PFI in presence of blur is calculated as
\begin{equation}
	\Psi=E\left[\psi\left(\mathbf{p}\right)\right]=K\int_{0}^{+\infty}{\left|G(\rho,\vartheta)\right|^2\left|B(\rho,\vartheta)\right|^2\rho d\rho}
\end{equation}
and, in the absence of blur,
\begin{equation}
	\widetilde{\Psi}=E\left[\widetilde{\psi}\left(\mathbf{p}\right)\right]=K\int_{0}^{+\infty}{\left|G(\rho,\vartheta)\right|^2\rho d\rho}
\end{equation}
where
\begin{equation}
	K=\frac{1}{\sigma_V^2}\int_{0}^{2\pi}{f(\vartheta) d\vartheta}
\end{equation}

From known results of integral calculus, we arrive at the following closed form results:
\begin{equation}
	\Psi=K\int_{0}^{+\infty}{e^{-2(s_G^2+s_B^2)\rho^2}\rho d\rho=}\frac{K}{4\left(s_G^2+s_B^2\right)}
	\label{eq:psi}
\end{equation}
\begin{equation}
	\widetilde{\Psi}=K\int_{0}^{+\infty}{e^{-2s_G^2\rho^2}\rho d\rho=}\frac{K}{4s_G^2}
	\label{eq:psitilde}
\end{equation}

Now, we first take it for granted that the perceived loss of positional uncertainty is proportional to the objective one, considering the consistency of the position estimates in the 3D Euclidean space during the interaction of the subject with its surrounding world. Then, according to classical psycho-physical laws (Weber-Fechner law \cite{DICLAUDIO21}), we assume that the differential sensitivity to stimuli is inversely proportional to the size of the initial stimuli.

Identifying the stimulus with the \emph{positional uncertainty} (see (\ref{eq:psi}) and (\ref{eq:psitilde})), the following index is then proposed as a psycho-physically grounded model of the quality loss perception caused by Gaussian blur applied to natural images:
\begin{IEEEeqnarray}{l}
	\varepsilon(\xi)=\frac{\sqrt{\displaystyle\frac{1}{\Psi}}-\sqrt{\displaystyle\frac{1}{\widetilde{\Psi}}}}{\sqrt{\displaystyle\frac{1}{\Psi}}}=\frac{\sqrt{\displaystyle s_G^2+s_B^2}-\sqrt{\displaystyle s_G^2}}{\sqrt{\displaystyle s_G^2+s_B^2}}= \nonumber\\
	=1-\frac{\sqrt{\displaystyle s_G^2}}{\sqrt{\displaystyle s_G^2+s_B^2}}=1-\sqrt{\frac{1}{1+\xi^2}}
\end{IEEEeqnarray}
having introduced the \emph{normalized blur} $\xi\overset{\Delta}{=}\displaystyle\frac{s_B}{s_G}$, which constitutes the metric of this quality loss estimator. This index is straightforward mapped onto the DMOS scale (which assigns the value 0 to the perfect quality, and 100 to the “worst” quality) by rescaling it in centesimal units:
\begin{equation}
	{\hat{d}(\xi)}=100\left[1-\sqrt{\frac{1}{1+\xi^2}}\right]\; .
\end{equation}

\begin{figure}[!ht]
	\centering
	\includegraphics[width=3.7in]{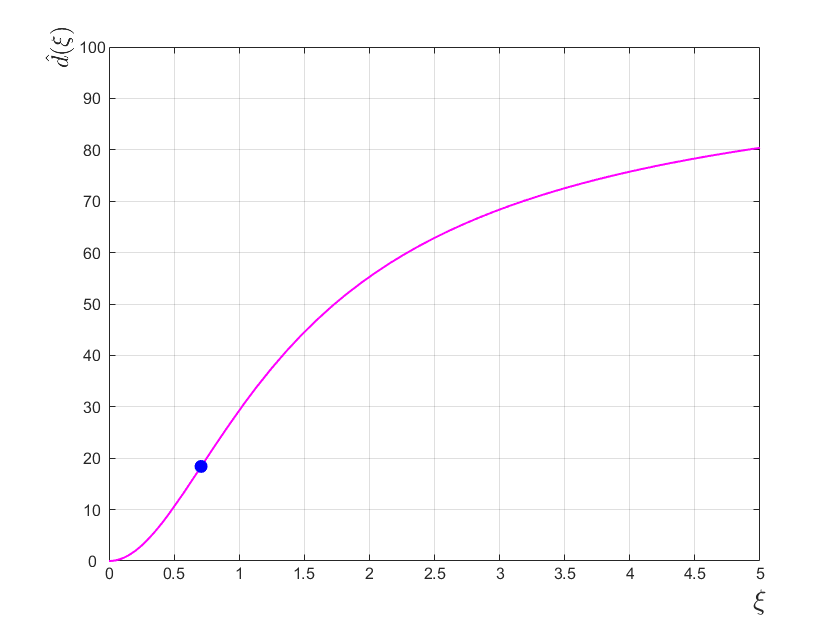}
	\caption{The rating function of the canonical DMOS estimator versus the metric $\xi$ (in centesimal units DMOS). The blue dot indicates the point of maximum sensitivity of the HVS with respect to the variations in blur.}
	\label{fig3}
\end{figure}

The shape of this theoretical DMOS estimator is plotted in Fig.\ref{fig3}. The curve exhibits a characteristic sigmoidal behavior, with a threshold for small input values and a saturation for large input values. The psycho-physical plausibility of this rating function is also supported by its compliance with the typical “dipper” shape \cite{SOLOMON09} of the incremental blur producing a Just Noticeable Difference (JND) of the perceived quality. Specifically, as shown in \cite{DICLAUDIO21}, its theoretical minimum occurs at
\begin{equation}
	\xi=\sqrt{\frac{1}{2}}
\end{equation}
corresponding to $\hat{d}(\xi)=18.4$ (see Fig.\ref{fig3}).

Still, this DMOS estimate remains undetermined. First, the concept of \emph{worst quality} is undefined. To fix it, it is necessary to assign by convention the subjective quality in correspondence of a chosen value of the blur spread. This is done by introducing a \emph{scoring gain} parameter $Q$ so that
\begin{equation}
	Q=\frac{d_A}{100}\left[1-\sqrt{\frac{1}{1+\xi_A}}\right]^{-1}
\end{equation}
where $d_A$ is a DMOS value conventionally attributed to a selected image for a certain value of $\xi_A$ (“anchor image”).

The above estimates refer to the nominal VD, i.e., to the case where $\delta=\delta_0$. Accounting for the scaling of the retinal image due to the actual visual distance we have
\begin{IEEEeqnarray}{l}
	\Psi=K\displaystyle\int_{0}^{+\infty}{e^{-2\left[\left(\displaystyle\frac{\delta}{\delta_0}s_G\right)^2+\left(\displaystyle\frac{\delta_0}{\delta}s_B\right)^2\right]\rho^2}\rho d\rho}= \nonumber\\
	=\frac{K}{4\left[\left(\displaystyle\frac{\delta}{\delta_0}s_G\right)^2+\left(\displaystyle\frac{\delta_0}{\delta}s_B\right)^2\right]} \nonumber\\
	\vspace{3mm}\widetilde{\Psi}=\displaystyle\frac{K}{4\left(\displaystyle\frac{\delta}{\delta_0}s_G\right)^2}
\end{IEEEeqnarray}

Therefore, defining the normalized VD $\tau\overset{\Delta}{=}\displaystyle\frac{\delta}{\delta_0}$ for notational simplicity, we finally obtain the following theoretical formula for the prediction of the DMOS of \emph{blurred natural images}:
\begin{equation}
	{\hat{d}}_{CAN}(Q,\tau,\xi)=100\times Q\left[1-\sqrt{\frac{1}{1+\displaystyle\frac{\xi^2}{\tau^4}}}\right]
	\label{eqn:dblur}
\end{equation}
referred to as \emph{canonical estimate} of the DMOS of Gaussian blurred natural images. It is characterized by the scoring gain $Q$ and the dimensionless parameter $\tau$.

Incidentally, if a set of empirical DMOS data is available for Gaussian blurred images both $Q$ and $\tau$ can be blindly estimated by regression, provided that the distribution of the spectra of such images contained in the database is well balanced, i.e., that the empirical values of the DMOS scatter symmetrically around their expected value.

\begin{figure}[!ht]
	\centering
	\includegraphics[width=4.0in]{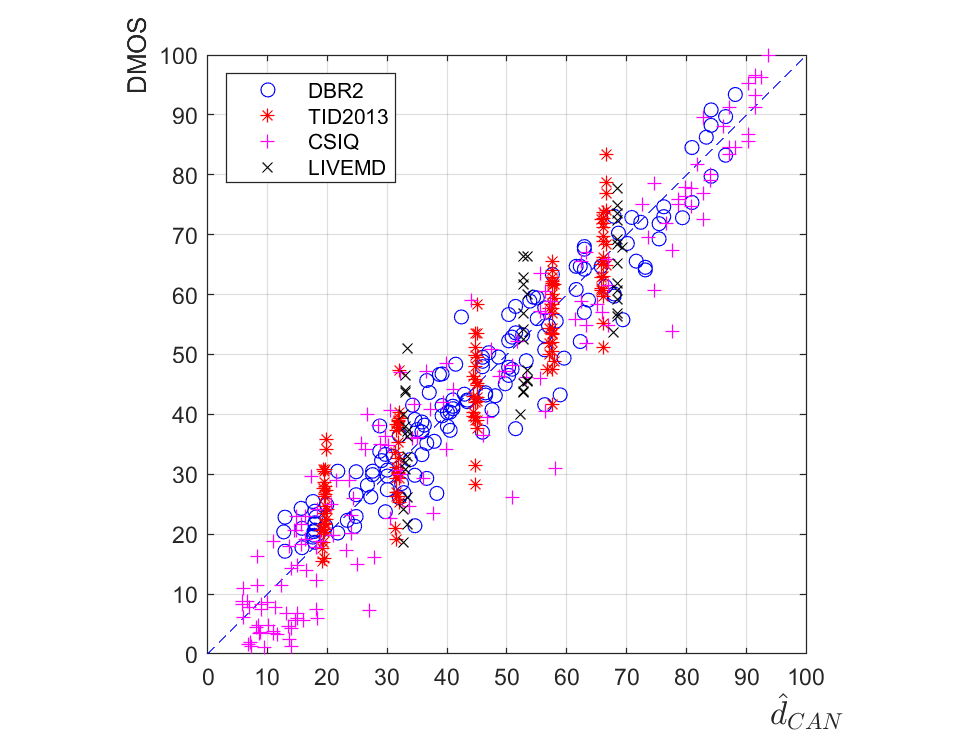}
	\caption{The scatterplots of the actual DMOS values versus the ones predicted by the canonical estimator for the IQA databases DBR2 (o), TID2013 (*),  CISQ (+), LIVE MD (x). These scatterplots corroborate the theoretically predicted linearity of the canonical estimator for natural images.}
	\label{fig18}
\end{figure}

Applying for instance this method to the blurred images of the LIVE Database Release 2 (DBR2) \cite{SHEIKH06B}, the Tampere Image Database (TID2013) \cite{PONOMARENKO15}, the LIVE Multi Distortion (MD) \cite{JAYARAMAN12}, and to the Computational and Subjective Image Quality Database (CSIQ) \cite{LARSON10}, it gives the scatterplots of the DMOS empirical values versus their estimates of Fig.\ref{fig18}. This scatterplot supports the supposed linearity of the canonical estimate with the empirical DMOS values. Here and in the following, the value of the metric $\xi$, i.e., of the spread $s_B$ is directly obtained from the blurred and reference images of the databases by straightforward spectral division calculated at different radial frequencies.

\section{Relationship among the canonical IQA method and other IQA methods}
\label{sec:Relationship among the canonical IQA method and other IQA methods}
\noindent The scope of this section is to analyze the possibility of linearizing a-priori the behavior of a generic IQA method using the properties of the canonical method. For the sake of clarity, let us limit preliminarily to the case of Gaussian blurred images.

Let us consider a generic IQA method based on the scalar metric $\zeta$, and let us assume that this metric is \emph{monotonically} related with the degree of Gaussian blur applied to images, i.e., with the canonical metric $\xi$.

Then it is possible, almost in principle, to convert the metric $\zeta$ into the metric $\xi$. Such a conversion is obtained by searching for the function $\xi(\zeta;\tau)$ that satisfies the equality of the DMOS estimates:
\begin{equation}
	{\hat{d}}_{CAN}(\xi;\tau)={\hat{d}}_{IQA}(\zeta;\tau)
\end{equation}

The function ${\hat{d}}_{CAN}(\xi;\tau)$ is available in an analytical closed form, whereas ${\hat{d}}_{IQA}(\zeta;\tau)$ is numerically calculable for a specific image. Therefore, to solve for the conversion function $\xi(\zeta;\tau)$ we adopted a semi-numerical method, resorting to a \emph{specimen} image ${\widetilde{I}}_{Typ}$ sufficiently representative of natural images, with respect to other factors influencing too the perceived quality loss, beyond Gaussian defocus and the related VD. First of all, the \emph{image content}.

Ideally, one could imagine synthesizing such an image \cite{MURRAY10}, but unfortunately, we do not have at present proper theoretical attributes of this image other than their spectral fall-off.

\begin{figure}[!ht]
	\centering
	\includegraphics[width=3.5in]{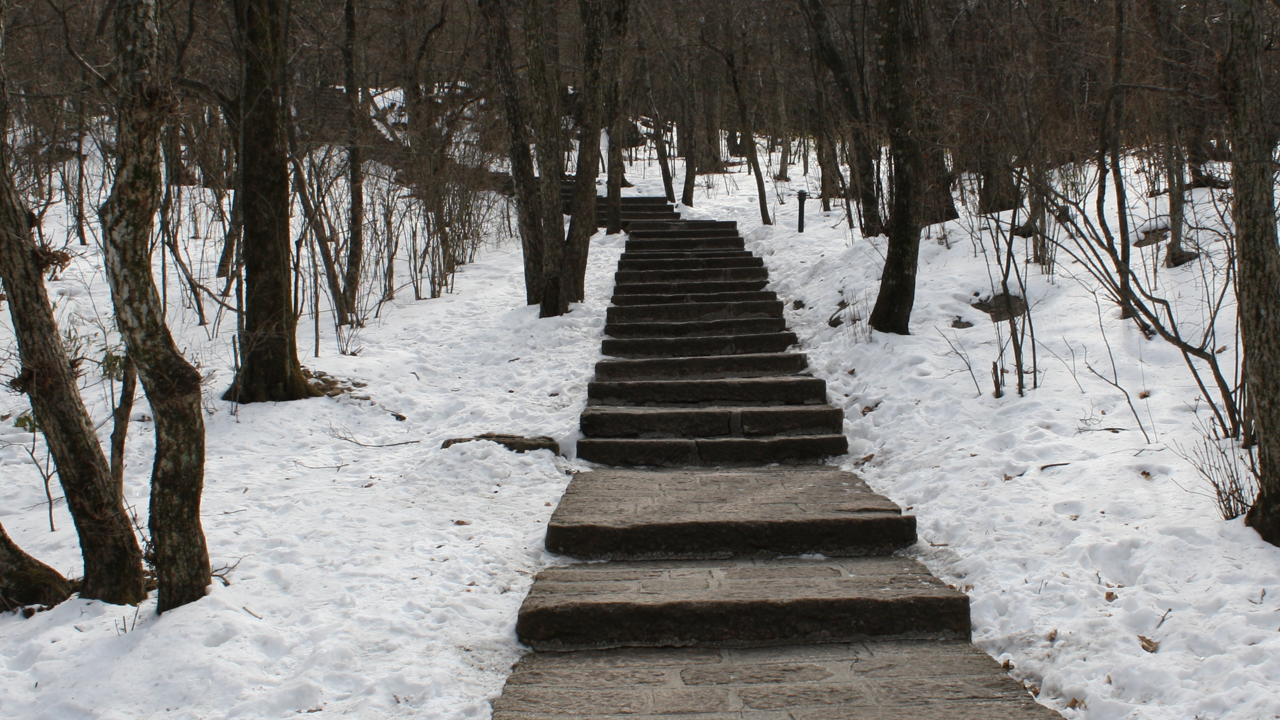}
	\caption{The specimen image “Iceroad” extracted from the database LIVE MD.}
	\label{fig5}
\end{figure}

The methods followed here is rather pragmatic. We searched for a specimen within one of the popular collections employed for statistical verification. Choosing the LIVE MD collection, whose images are \emph{not shared by the other collections}, and finally selected the image “Iceroad” (Fig.\ref{fig5}) as a performant specimen.

Subsequently a suite of fifty Gaussian blurred versions of this image spanning the whole DMOS range was generated. Based on these samples, the functions $\xi(\zeta;\tau)$ for different values of the VD were defined using a Piecewise Cubic Hermite Interpolating Polynomial (PCHIP) Matlab$\textsuperscript{\textregistered}$ function VDs. An example for $\tau=0.76$ is displayed in Fig.\ref{fig6} \footnote{The whole set of functions is available to the reader in the supplementary data.\label{footnote_1}}.

It is opportune to underline here that, strictly speaking, such a pragmatic procedure violates the immunity to empirical data, since it passes through an experimental verification phase. Nevertheless, it preserves the claimed calibration-free nature, since the functions $\xi(\zeta;\tau)$ are fixed once for all. These functions reveal unexplored aspects of  metrics.

Let us observe that the fundamental behavior of the function $\xi(\zeta;\tau)$ is analogous for all the considered methods. It compresses the equivalent blur values for highly degraded images and conversely expands them for slightly degraded images. In Fig.\ref{fig6a}, it is shown how the functions change with the VD in the VIF case.

\begin{figure}[!ht]
	\centering
	\includegraphics[width=3.7in]{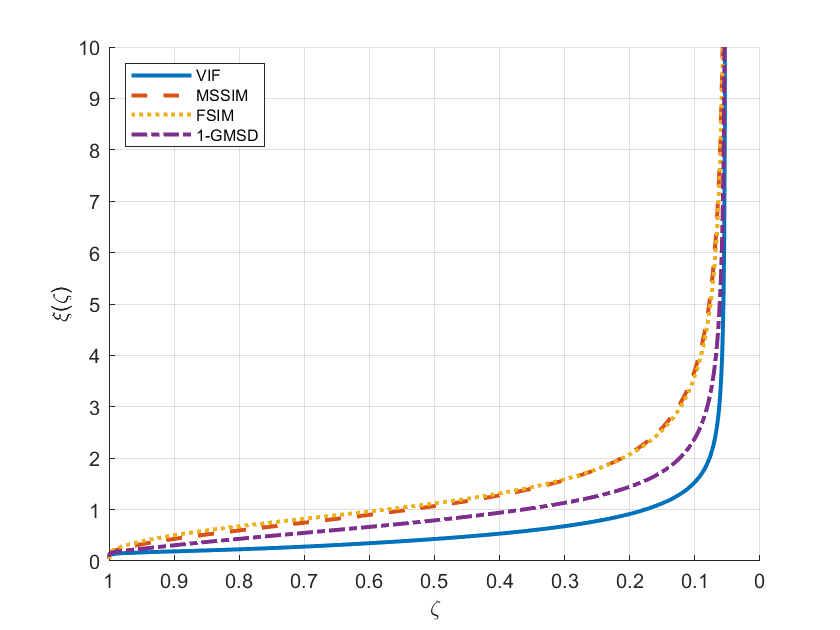}
	\caption{The metric conversion function for the considered IQA methods. The scales of the metrics of the MSSIM, FSIM and GMSD, for $\tau=0.76$, are purposely modified and aligned to the scale of the VIF metric to put into evidence their common asymptotic behavior.}
	\label{fig6}
\end{figure}

\begin{figure}[!ht]
	\centering
	\includegraphics[width=3.7in]{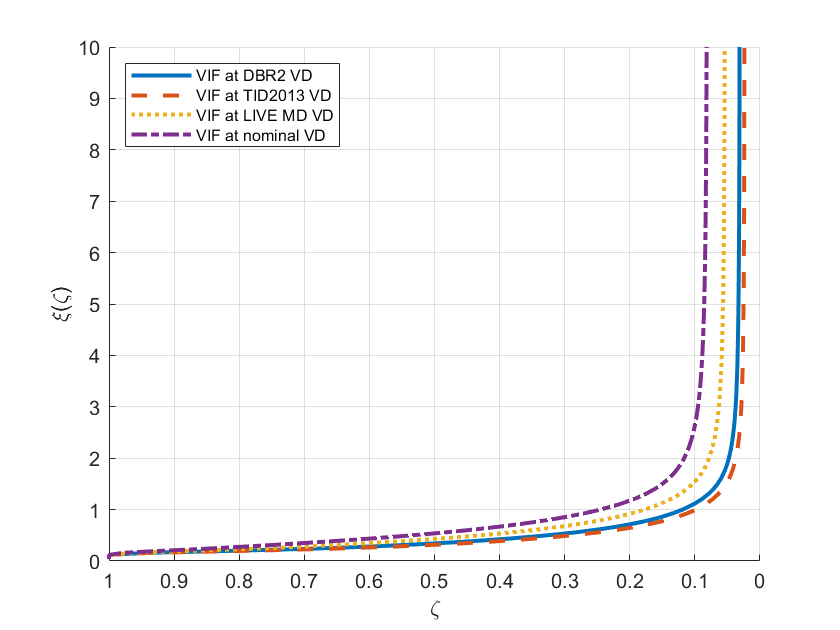}
	\caption{The metric conversion function for the VIF method at different VDs: $\tau=0.53$, $\tau=0.44$, $\tau=0.60$, and $\tau=0.76$, corresponding respectively to the LIVE DBR2, TID2013, CSIQ and LIVE MD VDs, and $\tau=1$ for the nominal VD.}
	\label{fig6a}
\end{figure}

Once mapped onto the equivalent Gaussian blur axis $\xi$, a linear estimate $\hat{d}(\zeta)$ of the DMOS Gaussian of blurred natural images, is finally given by the CRF
\begin{equation}
	{\hat{d}}(\zeta)=100\times Q\left[1-\sqrt{\frac{1}{1+\displaystyle\frac{\xi(\zeta;\tau)^2}{\tau^4}}}\right]\; .
	\label{eq:dest_zeta}
\end{equation}

\begin{figure}[!ht]
	\centering
	\includegraphics[width=3.85in]{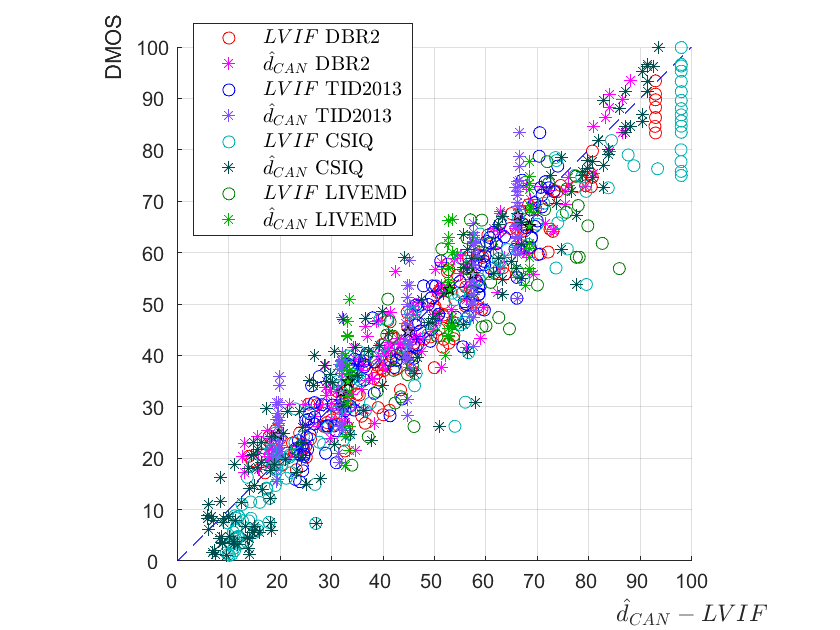}
	\caption{The scatterplots of the DMOS empirical measurements  for blurred images versus the values predicted using respectively the canonical method (*) and the linearized VIF method (°) for different databases. The vertical alignments of the DMOS values in the cases of the TID2013 (purple stars) and LIVE MD (green stars) databases are destroyed by the metric conversion (purple and green circles).}
	\label{fig7}
\end{figure}

To give a visual account of the linearity and of the statistical accuracy of this IQA method, in Fig.\ref{fig7} the scatterplots of the DMOS estimates for the canonical method against the same estimates obtained with the calibration-free VIF method are provided for Gaussian blur. Notice, in particular, that the vertical alignments of the DMOS values in the cases of the TID2013 and LIVE MD databases due to discrete values of $s_B$ are destroyed by the metric conversion.

\section{Extension of the method beyond Gaussian blur}
\label{sec:Extension of the method beyond Gaussian blur}
\noindent The analysis conducted so far, limited to the case of Gaussian blurred natural images, is of limited practical interest. From an applicative viewpoint, it is important to have a method for assessing the quality of images subject to different types of degradation.

At a glance, it could appear that the conversion to the metric $\xi$ of a generic IQA metric $\zeta$ is vane, since the metric $\xi$ measures only one type of degradation, that cannot reflect for instance the effect of noise, which is neatly distinct with respect to blur from a perceptual viewpoint \cite{MARTENS96}.

However, the metrics of some mainstream methods are purposely designed to provide measures well correlated with the human quality score, that would produce the same MOS/DMOS even if its causes are \emph{different} \cite{WANG09}. This aspect is explicitly discussed in \cite{SHEIKH06} where the concept of perceptual quality equivalence for some degradation viewed as mixtures of noise and blur is introduced, and in \cite{DICLAUDIO18} where two separate metrics are jointly employed and a two-dimensional scoring function is defined. Therein, it was outlined that the perceptual quality equivalence among blur and noise is \emph{interpersonal} and then, assumed as constant. See also \cite{CARON11}.

This implies that the conversion function $\xi(\zeta;\tau)$ makes sense even if the values of the metric $\zeta$ are not only determined by the presence of blur, but by other factors that produce the same values of  ${\hat{d}}_{CAN}(\xi)={\hat{d}}_{IQA}(\zeta)$. This is specifically the case of the considered metrics. For instance, in the VIF method the metric is computed as the contribution of many terms measuring losses of Shannon information, where the effect of increasing noise is equivalent to that caused by the reduction of the image signal energy due to blur.

The methods obtained in this way are herein referred to as \emph{Linearized IQA} (LIQA), whose estimates have the same form as the (\ref{eq:dest_zeta}):
\begin{equation}
	{\hat{d}}_{LIQA}(\zeta)=100\times Q\left[1-\sqrt{\frac{1}{1+\displaystyle\frac{\xi(\zeta;\tau)^2}{\tau^4}}}\right]\; .
\end{equation}

\section{Performance evaluation}
\label{sec:Performance evaluation}
\noindent The expected operational advantage of the calibration-free procedure is to allow comparison of MOS/DMOS estimates across different applications and VDs. The expected toll is a lower accuracy within contexts characterized by possibly unmodeled causes of quality loss.

To tangibly guess the entity of this trade-off, we present here a statistical performance comparison of some popular classical IQA methods already considered above, subject to final calibration according to VQEG recommendations, with their LIQA counterparts, which do not make use of empirical calibration.

The present comparative analysis is limited to degradation types that appeared suitable for conversion into the canonical metric, namely, additive noise, JPEG and JPEG2000 compression, including some examples of \emph{multiple degradations} \cite{JAYARAMAN12}. To this purpose, four different well-known collections of examples have been employed, namely the above indicated subsets of the LIVE DBR2, the TID2013, the CISQ and the LIVE MD databases. These databases are built using independent protocols and methods for calculating the DMOS, which is used here as a common measure of quality loss.

\begin{table*}[!ht]
	\centering
	\caption{Experimental verification for the \textbf{a-priori linearized IQA metrics} (left table) and \textbf{empirically calibrated IQA metrics} (right table) for LIVE DBR2, TID2013, CSIQ and LIVE MD. Four distortions subsets for LIVE DBR2, TID2013 and CSIQ: blur, noise, JPEG, JPEG2000. Five distortions subsets for LIVE MD: blur, noise, JPEG, blur+JPEG, blur+noise. The VDs employed in the different databases were inferred by regression using the quality estimates for the subsets of blurred images, giving the following normalized values: $\tau=0.53$ for LIVE DBR2, $\tau=0.44$ for TID2013, $\tau=0.60$ for CSIQ, and $\tau=0.76$ for LIVE MD.}
	\begin{tabular}{ cc }
		\scalebox{1}{
			\begin{tabular}{lccccc}
				\noalign{\smallskip}
				\toprule
				Model & RMSE & SROCC & LCC \\
				\midrule
				LVIF LIVE DBR2 & $8.6653$ & $0.96536$ & $0.95636$ \\
				LVIF TID2013 & $7.0170$ & $0.92782$ & $0.92789$ \\
				LVIF CSIQ & $7.5789$ & $0.95878$ & $0.96336$ \\
				LVIF LIVE MD & $9.2065$ & $0.92673$ & $0.93604$ \\
				\noalign{\smallskip}
				\hline
				\noalign{\smallskip}
				LMSSIM LIVE DBR2 & $9.5823$ & $0.95323$ & $0.93910$ \\
				LMSSIM TID2013 & $7.8579$ & $0.91745$ & $0.91394$ \\
				LMSSIM CSIQ & $8.5463$ & $0.95445$ & $0.95326$ \\
				LMSSIM LIVE MD & $13.8323$ & $0.87709$ & $0.84556$ \\
				\noalign{\smallskip}
				\hline
				\noalign{\smallskip}
				LFSIM LIVE DBR2 & $8.8302$ & $0.96851$ & $0.95149$ \\
				LFSIM TID2013 & $7.3283$ & $0.95341$ & $0.93291$ \\
				LFSIM CSIQ & $8.1802$ & $0.96164$ & $0.95810$ \\
				LFSIM LIVE MD & $12.0365$ & $0.91389$ & $0.88788$ \\
				\noalign{\smallskip}
				\hline
				\noalign{\smallskip}
				LGMSD LIVE DBR2 & $7.2629$ & $0.96888$ & $0.96478$ \\
				LGMSD TID2013 & $6.7091$ & $0.95059$ & $0.93579$ \\
				LGMSD CSIQ & $7.7122$ & $0.96909$ & $0.96232$ \\
				LGMSD LIVE MD & $12.9031$ & $0.90144$ & $0.87629$ \\
				\bottomrule
			\end{tabular}
		}
		\hspace{20mm}
		\scalebox{1}{
			\begin{tabular}{lccccc}
				\noalign{\smallskip}
				\toprule
				Model & RMSE & SROCC & LCC \\
				\midrule
				VIF LIVE DBR2 & $7.3992$ & $0.96535$ & $0.96181$ \\
				VIF TID2013 & $6.3344$ & $0.92782$ & $0.94021$ \\
				VIF CSIQ & $9.5763$ & $0.92283$ & $0.93089$ \\
				VIF LIVE MD & $7.3096$ & $0.92199$ & $0.94277$ \\
				\noalign{\smallskip}
				\hline
				\noalign{\smallskip}
				MSSIM LIVE DBR2 & $8.4663$ & $0.95307$ & $0.94969$ \\
				MSSIM TID2013 & $6.7297$ & $0.91745$ & $0.93224$ \\
				MSSIM CSIQ & $13.1528$ & $0.88089$ & $0.86502$ \\
				MSSIM LIVE MD & $10.0285$ & $0.87118$ & $0.88922$ \\
				\noalign{\smallskip}
				\hline
				\noalign{\smallskip}
				FSIM LIVE DBR2 & $7.2289$ & $0.96846$ & $0.96358$ \\
				FSIM TID2013 & $5.4113$ & $0.95343$ & $0.95673$ \\
				FSIM CSIQ & $10.5632$ & $0.92742$ & $0.91522$ \\
				FSIM LIVE MD & $7.9413$ & $0.90831$ & $0.93207$ \\
				\noalign{\smallskip}
				\hline
				\noalign{\smallskip}
				GMSD LIVE DBR2 & $6.9278$ & $0.96888$ & $0.96660$ \\
				GMSD TID2013 & $5.5523$ & $0.95059$ & $0.95440$ \\
				GMSD CSIQ & $8.1502$ & $0.95490$ & $0.95044$ \\
				GMSD LIVE MD & $8.7612$ & $0.89582$ & $0.91666$ \\
				\bottomrule
			\end{tabular}
		}
	\end{tabular} \\
	\label{table:rectified-original-IQA}
\end{table*}

Specifically, in the LIVE DBR2 experiments used single stimulus based judgments, including reference images \cite{ITU12B}. Subjects were asked to express a score among five quality levels with a slider.

In the TID2013 experiments, a tristimulus methodology \cite{PONOMARENKO15} was adopted, where subjects simply selected the best between two degraded images in the presence of the reference image.

In the CSIQ experiments, subjects were asked to order the quality of four images simultaneously displayed on an array of monitors \cite{LARSON10}.

In the LIVE MD, a single stimulus \cite{ITU12B} with hidden reference methodology was adopted.

These basic differences, along with diverse viewing conditions and protocols, ensure that the selected series of measurements, if they yield coherent results, will reflect the essential nature of the subjective quality rating phenomenon, independent of the way it was observed and measured.

Even if indications about the VD used in experiments are available, both the normalized VD $\tau$ and the scoring gain $Q$ were independently estimated by regression of the available blur DMOS data, using the canonical estimator \cite{DICLAUDIO21}.

The results of this comparative analysis are reported in Tab.\ref{table:rectified-original-IQA} (left table) for the IQA methods after empirical calibration \cite{VQEG00,VQEG03}, and in the right table for the calibration-free LIQA methods. The selected statistical quality indices were the RMSE (indicative of the average distance between the actual DMOS values and the ones provided by the method), the SROCC (Spearman Rank Correlation Coefficient) which reveals the monotonicity between predicted and actual DMOS values, and the PLCC (Pearson Linear Correlation Coefficient) which measures the linearity of the DMOS mapping.

These tables indicate that the DMOS estimates based on theoretical modeling are statistically slightly more dispersed than the ones determined using empirical fitting to specific data, using the five parameter curve of (\ref{eq:VQEG}). However, this advantage does depend on the adopted fitting model. For instance, the three-parameter curve of \cite{ITU16} would exhibit in principle a larger RMSE in change of a better stability across applications.

On the other hand, the PLCC and the SROCC indices remain substantially the same for the IQAs and LIQAs methods.

\begin{figure*}[!htb]
	\centering
	\includegraphics[width=1.7in]{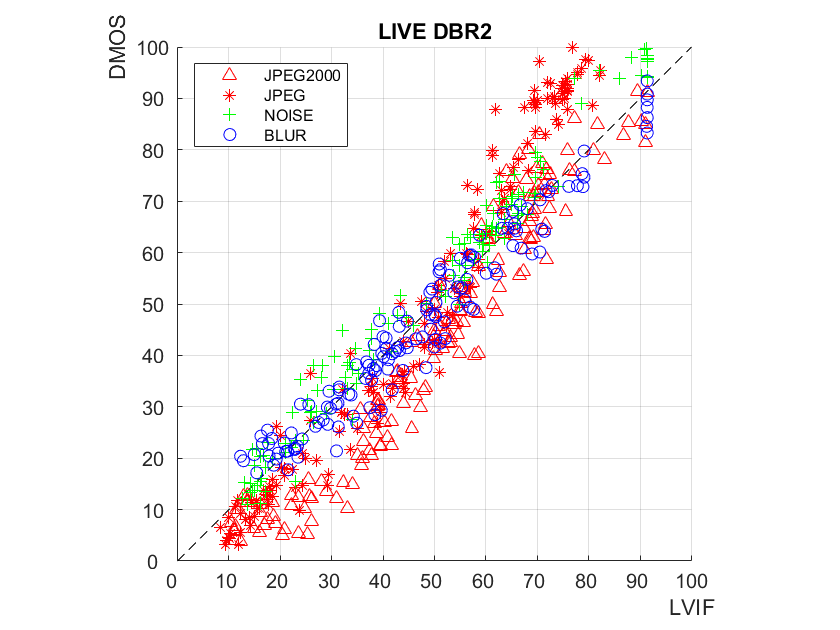}
	\includegraphics[width=1.7in]{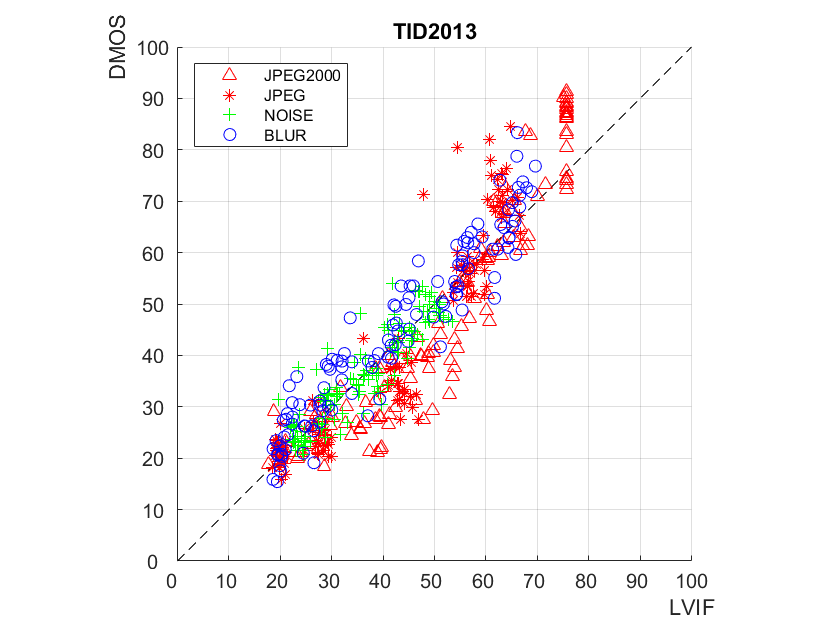}
	\includegraphics[width=1.7in]{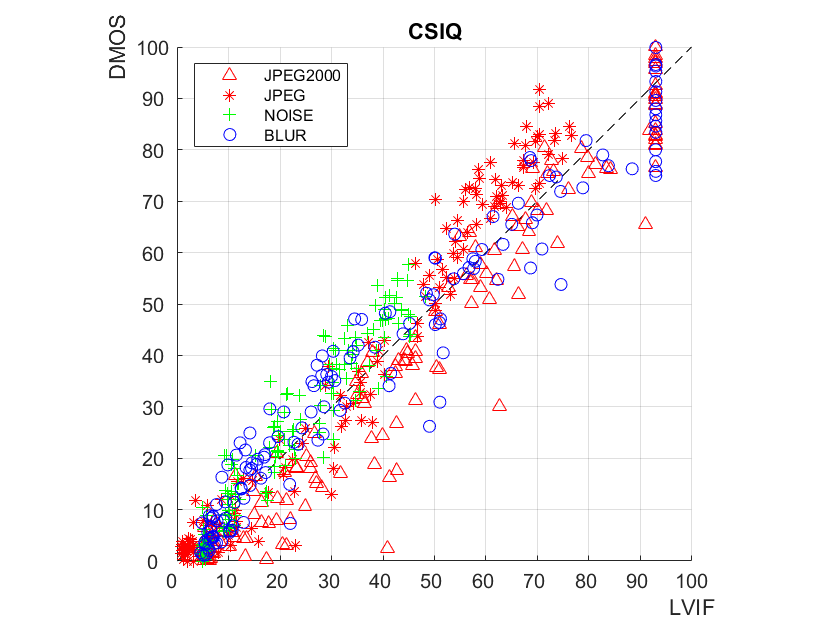}
	\includegraphics[width=1.7in]{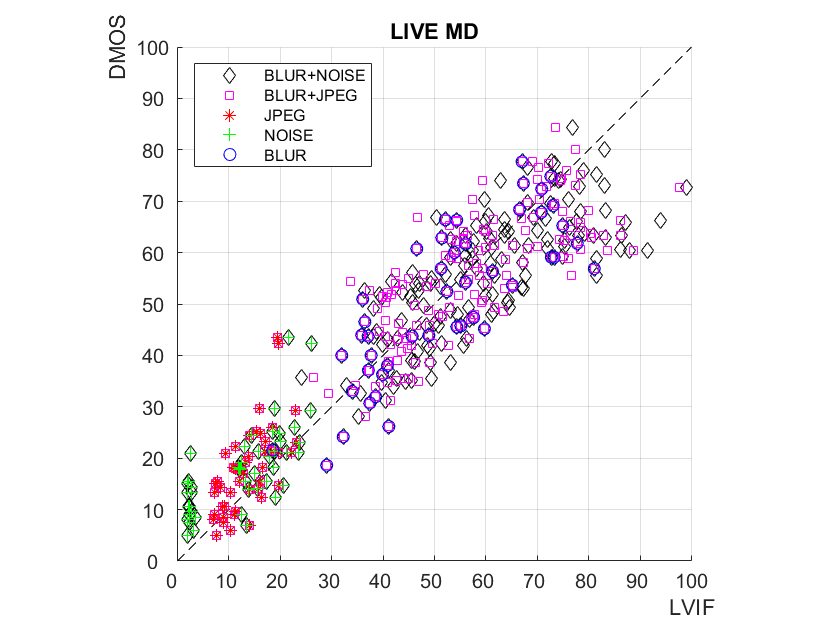}
	\includegraphics[width=1.7in]{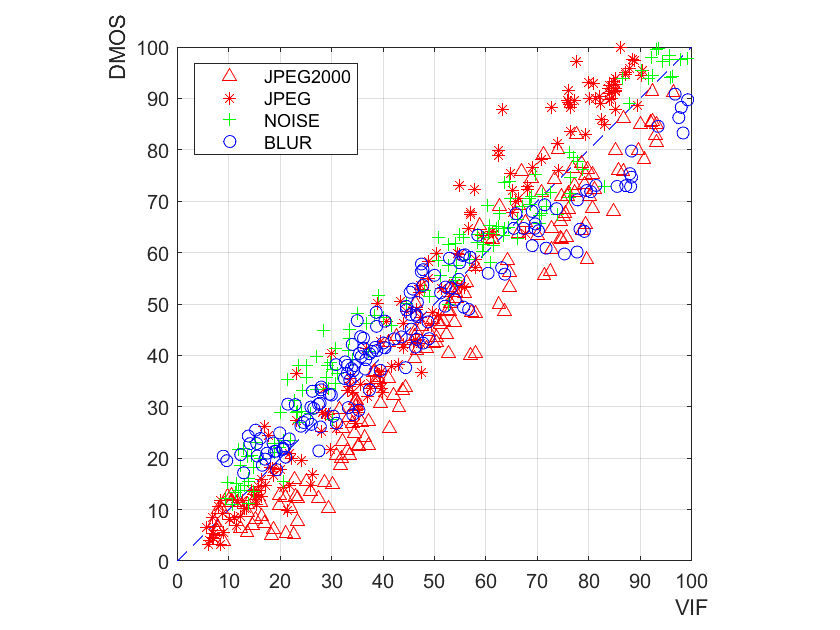}
	\includegraphics[width=1.7in]{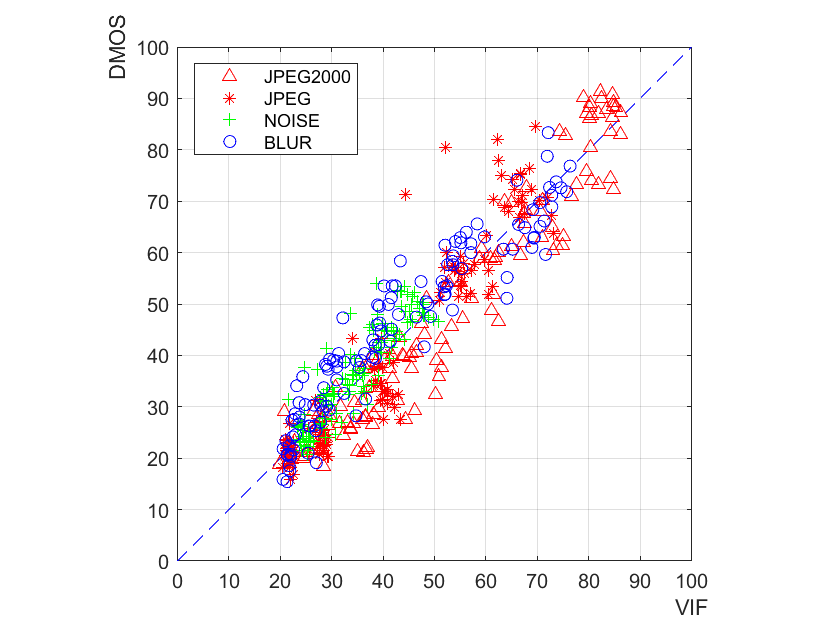}
	\includegraphics[width=1.7in]{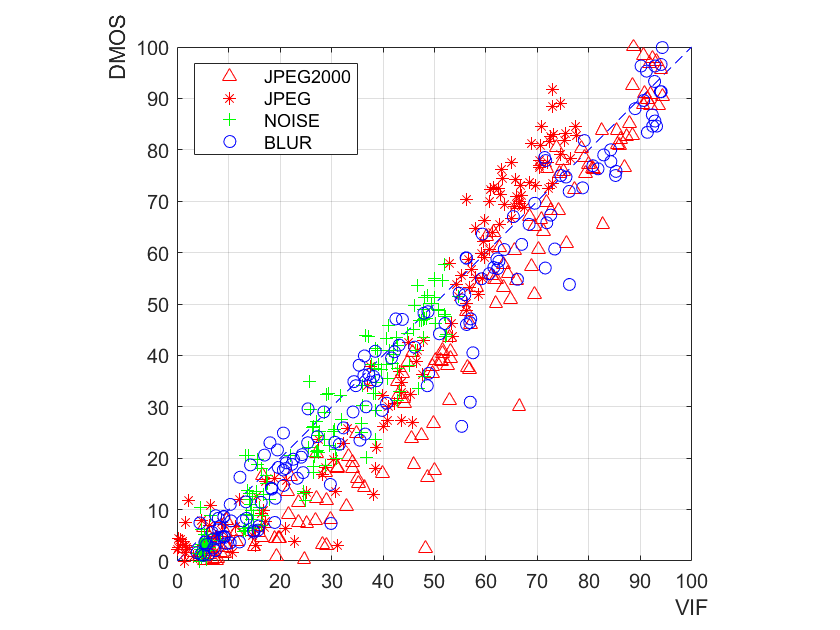}
	\includegraphics[width=1.7in]{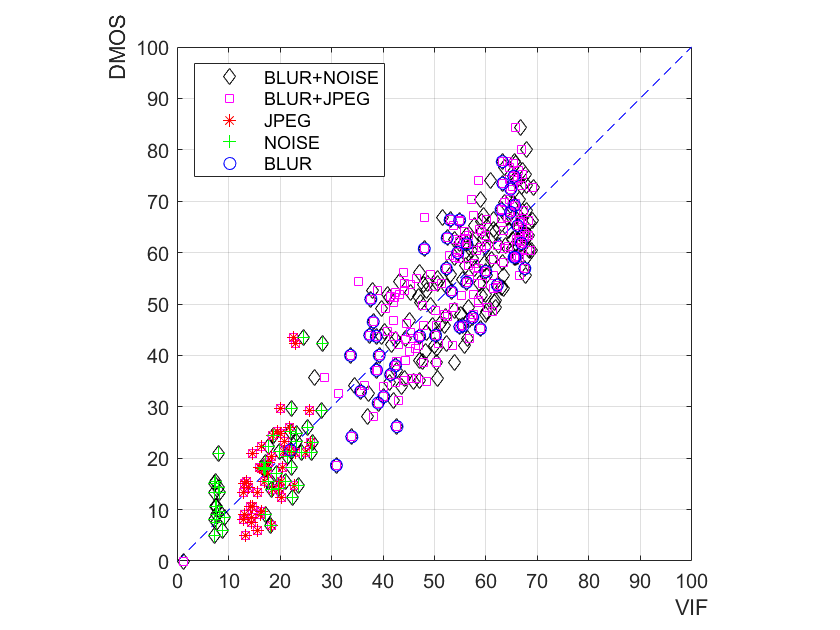}
	\caption{The DMOS scatterplots after empirical calibration for blur, noise, JPEG, JPEG2000 distortions for LIVE DBR2, TID2013 and CSIQ, and blur, noise, JPEG, blur+JPEG, blur+noise distortions for LIVE MD, versus the values predicted using respectively the theoretically linearized VIF method (upper row) and the conventional, empirically calibrated VIF method (lower row) for different databases.}
	\label{fig8}
\end{figure*}

\begin{figure*}[!htb]
	\centering
	\includegraphics[width=1.7in]{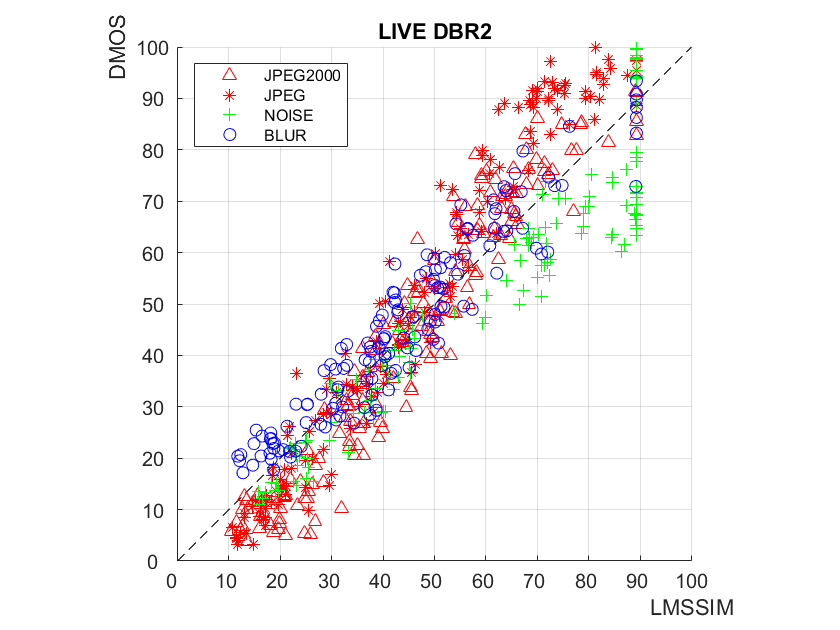}
	\includegraphics[width=1.7in]{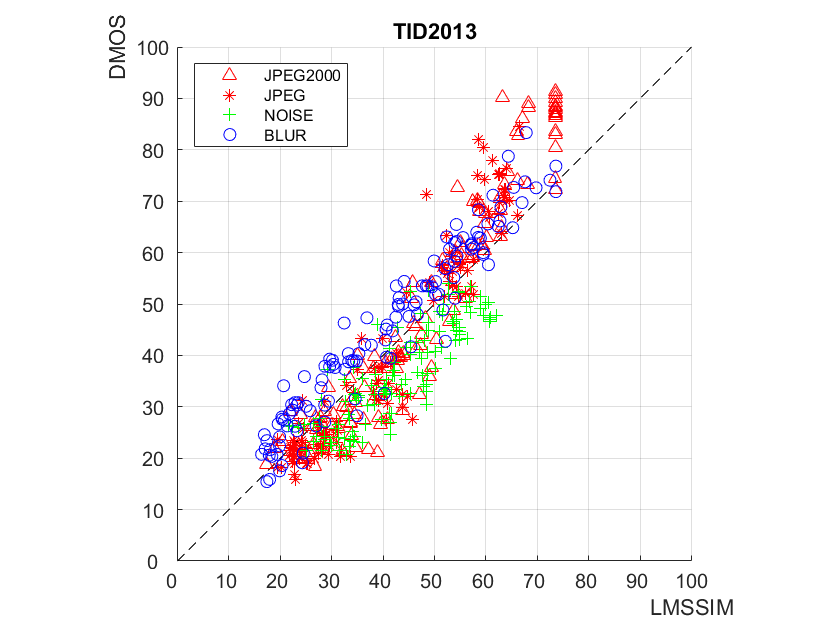}
	\includegraphics[width=1.7in]{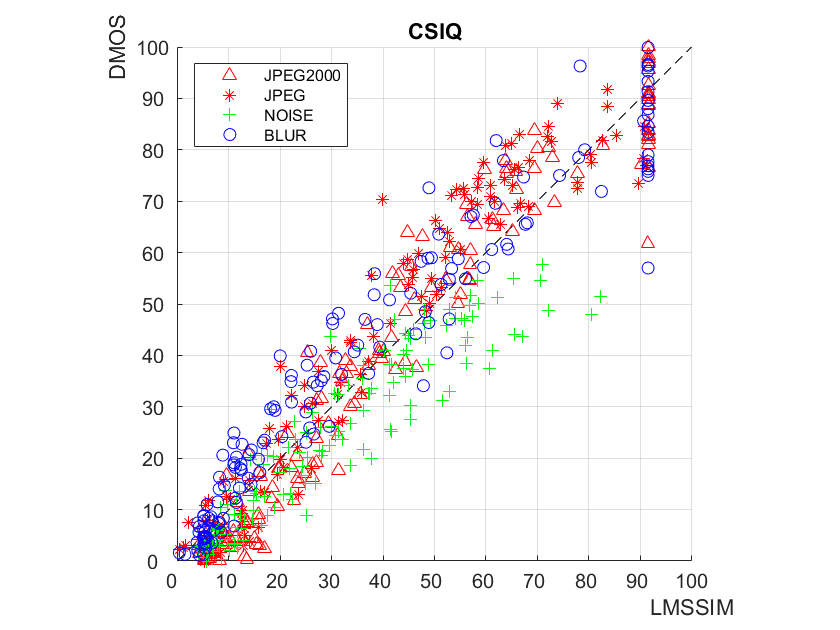}
	\includegraphics[width=1.7in]{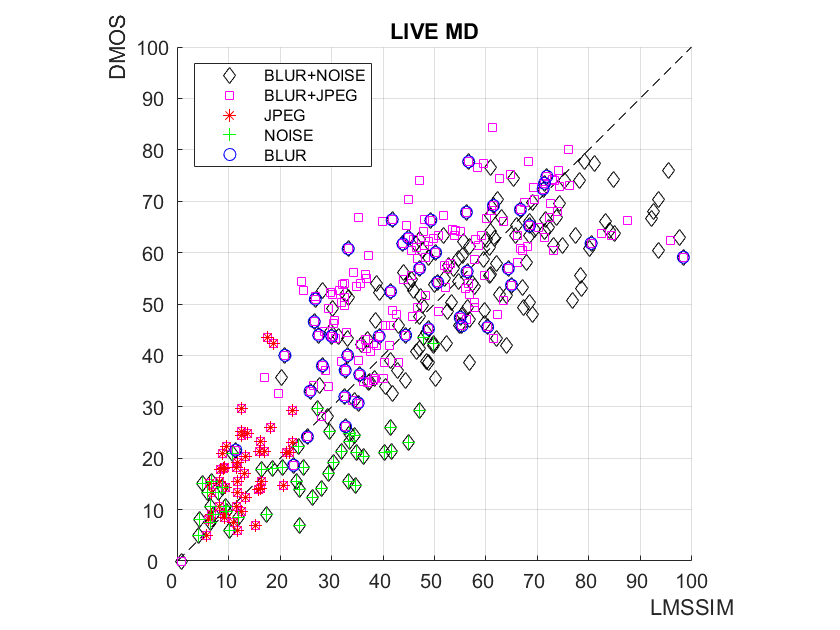}
	\includegraphics[width=1.7in]{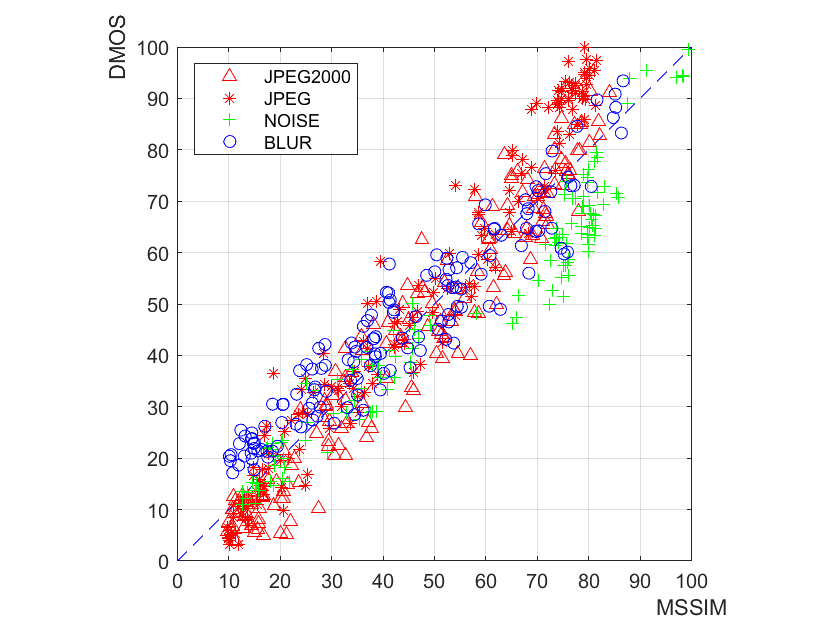}
	\includegraphics[width=1.7in]{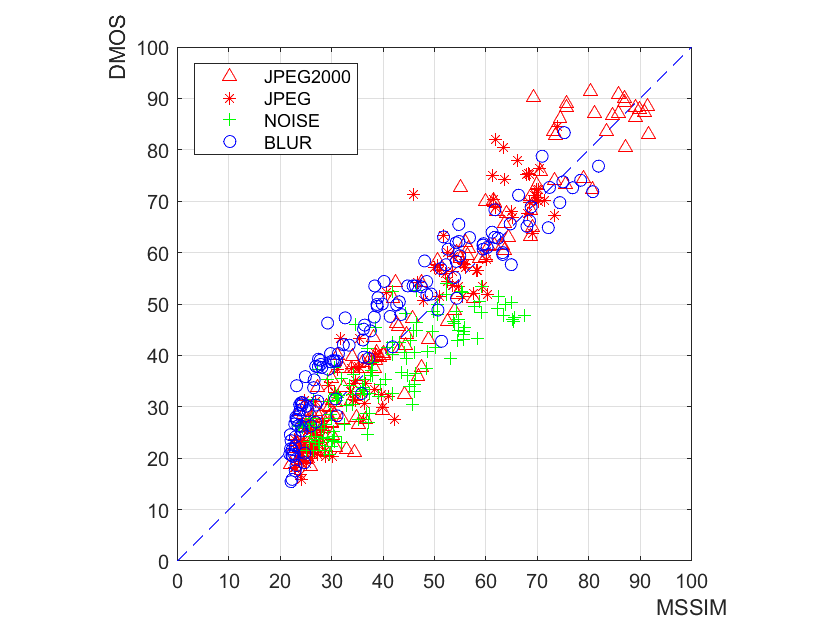}
	\includegraphics[width=1.7in]{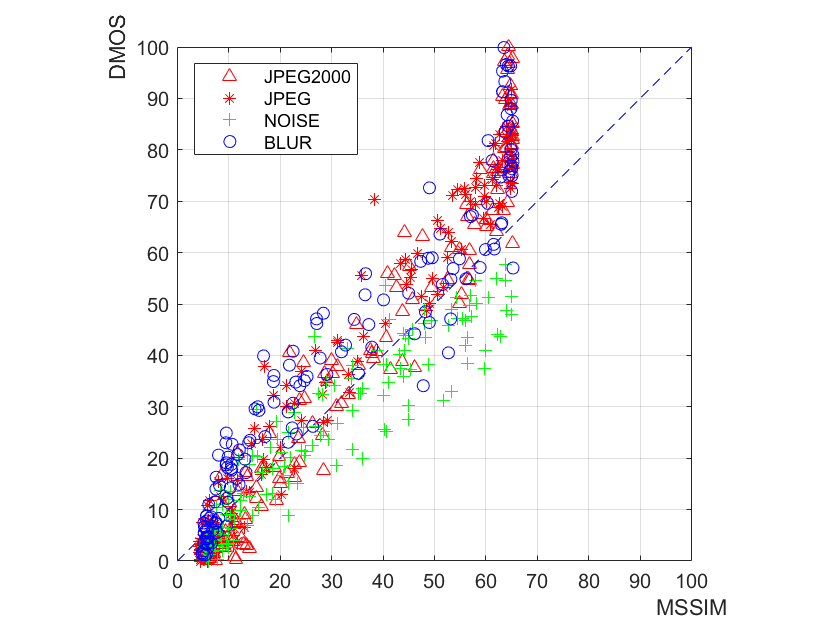}
	\includegraphics[width=1.7in]{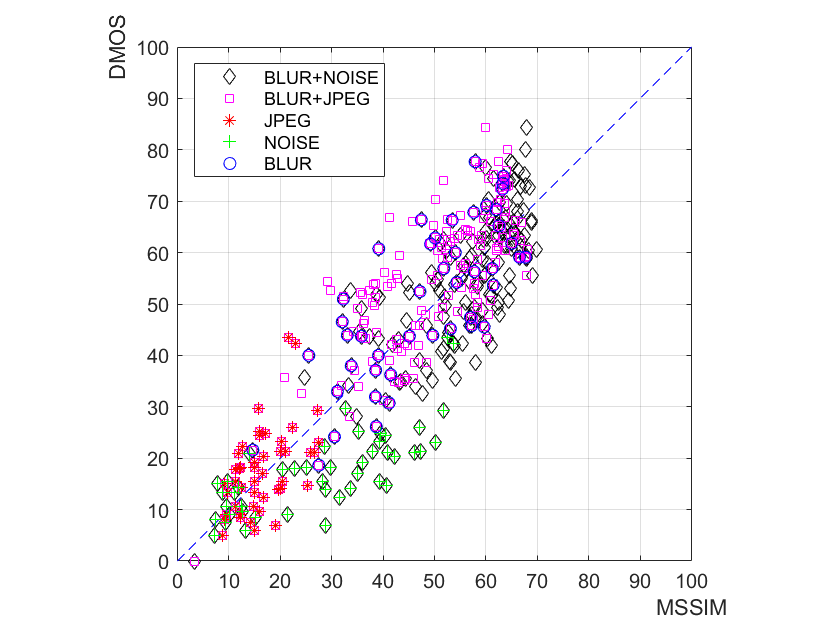}
	\caption{The DMOS scatterplots after empirical calibration for blur, noise, JPEG, JPEG2000 distortions for LIVE DBR2, TID2013 and CSIQ, and blur, noise, JPEG, blur+JPEG, blur+noise distortions for LIVE MD, versus the values predicted using respectively the theoretically linearized MSSIM method (upper row) and the conventional, empirically calibrated MSSIM method (lower row) for different databases.}
	\label{fig9}
\end{figure*}

\begin{figure*}[!htb]
	\centering
	\includegraphics[width=1.7in]{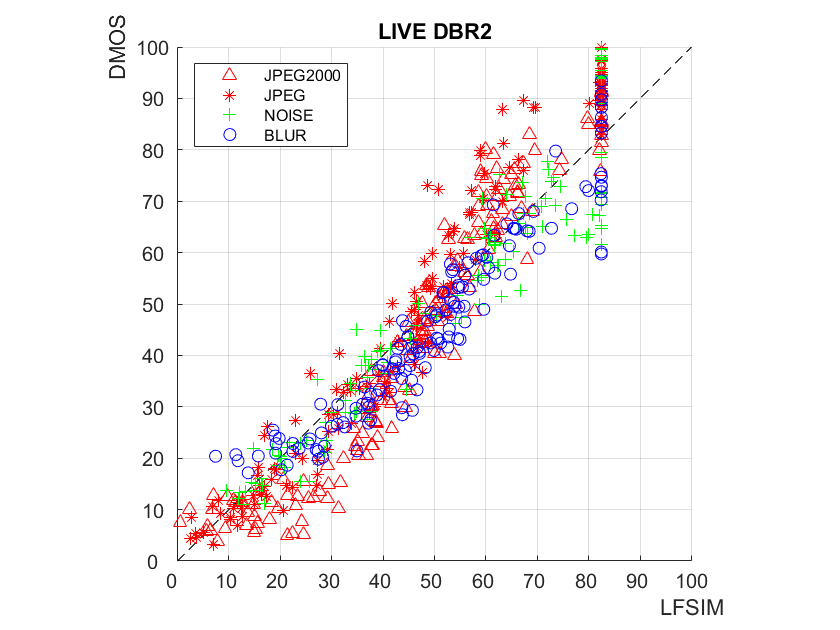}
	\includegraphics[width=1.7in]{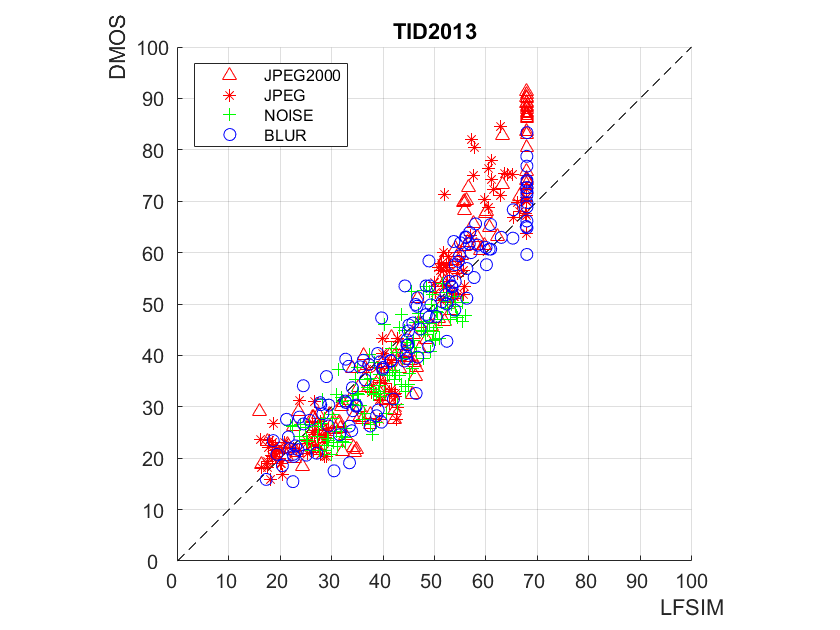}
	\includegraphics[width=1.7in]{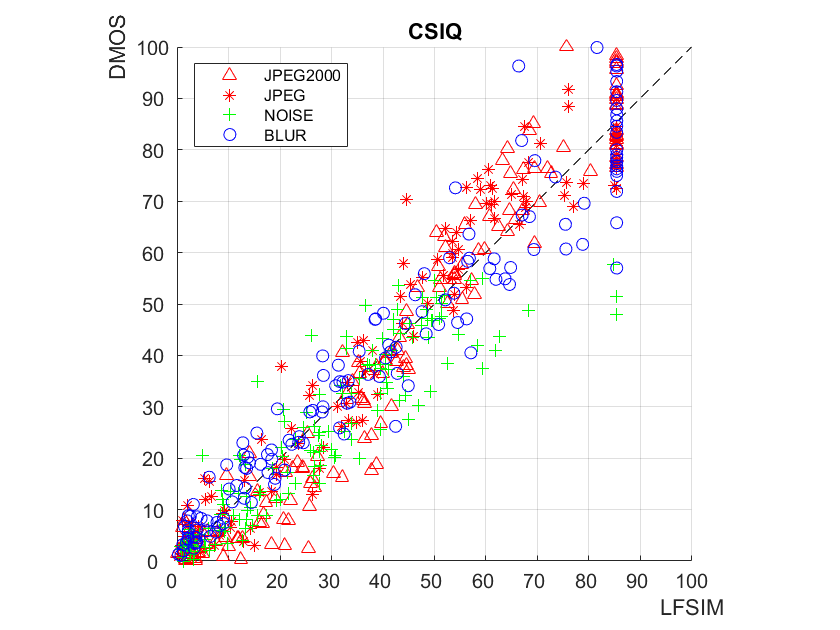}
	\includegraphics[width=1.7in]{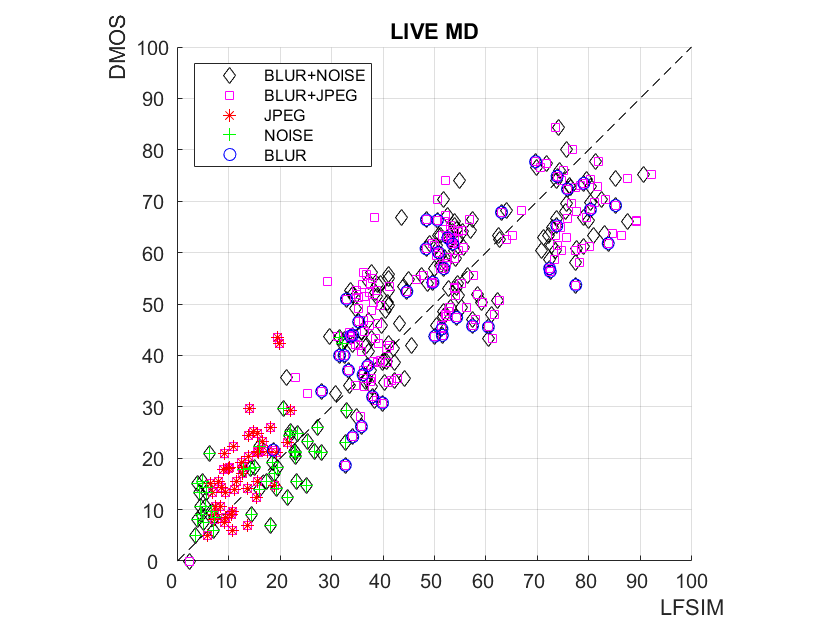}
	\includegraphics[width=1.7in]{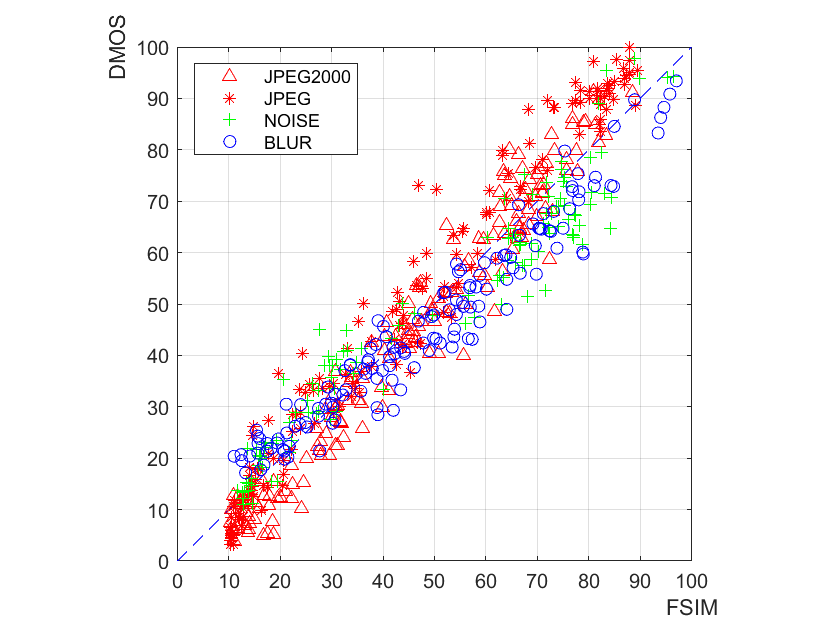}
	\includegraphics[width=1.7in]{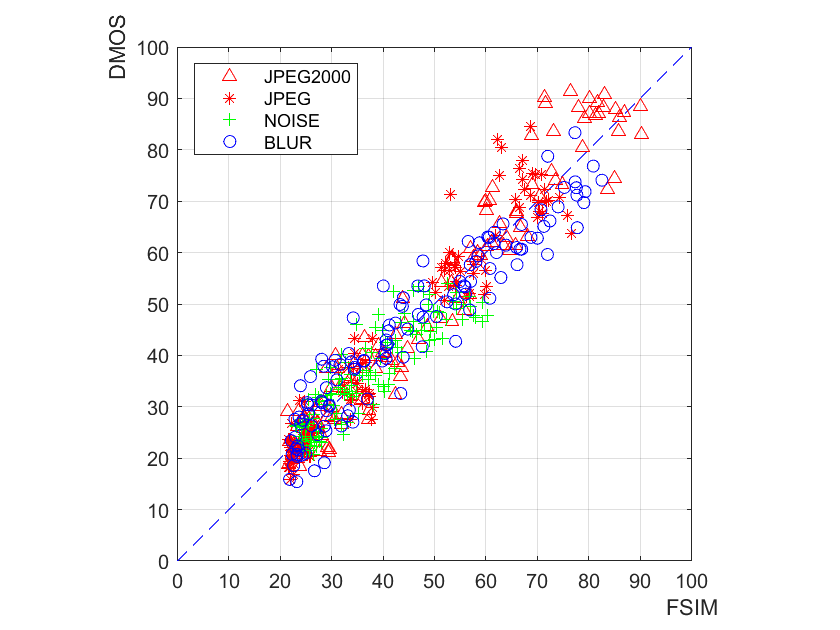}
	\includegraphics[width=1.7in]{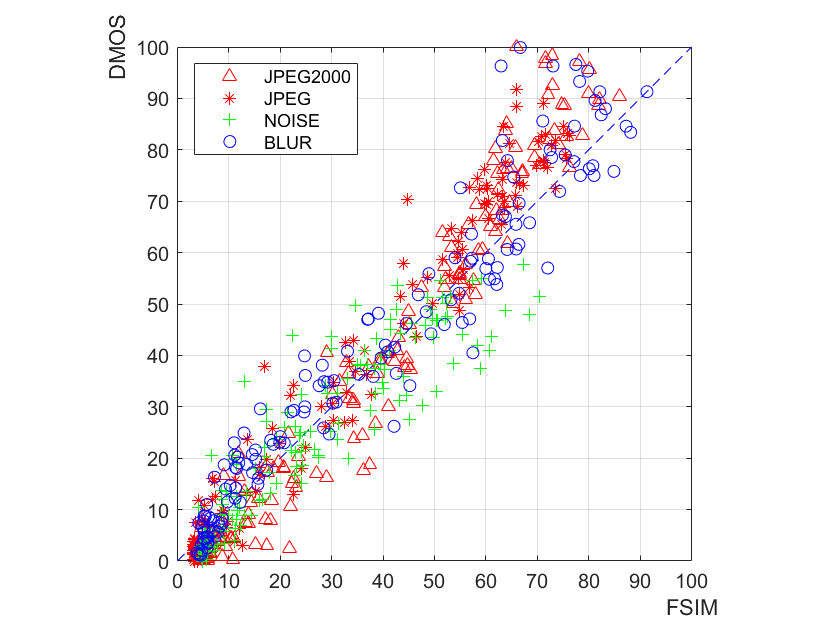}
	\includegraphics[width=1.7in]{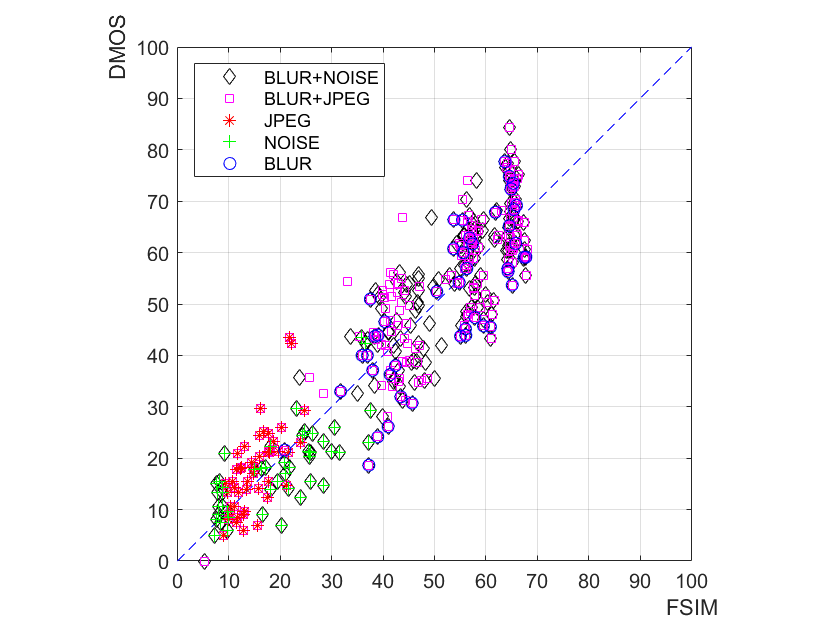}
	\caption{The DMOS scatterplots after empirical calibration for blur, noise, JPEG, JPEG2000 distortions for LIVE DBR2, TID2013 and CSIQ, and blur, noise, JPEG, blur+JPEG, blur+noise distortions for LIVE MD, versus the values predicted using respectively the theoretically linearized FSIM method (upper row) and the conventional, empirically calibrated FSIM method (lower row) for different databases.}
	\label{fig10}
\end{figure*}

\begin{figure*}[!htb]
	\centering
	\includegraphics[width=1.7in]{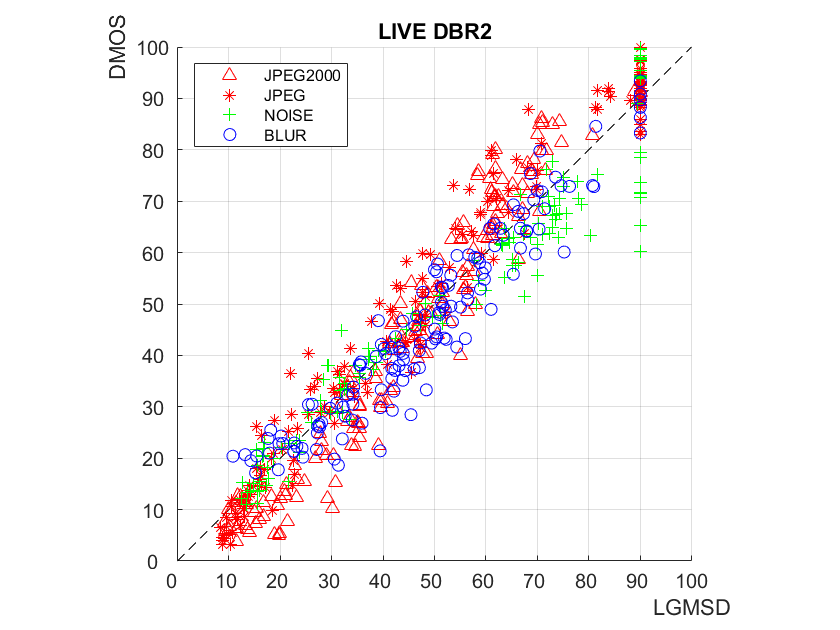}
	\includegraphics[width=1.7in]{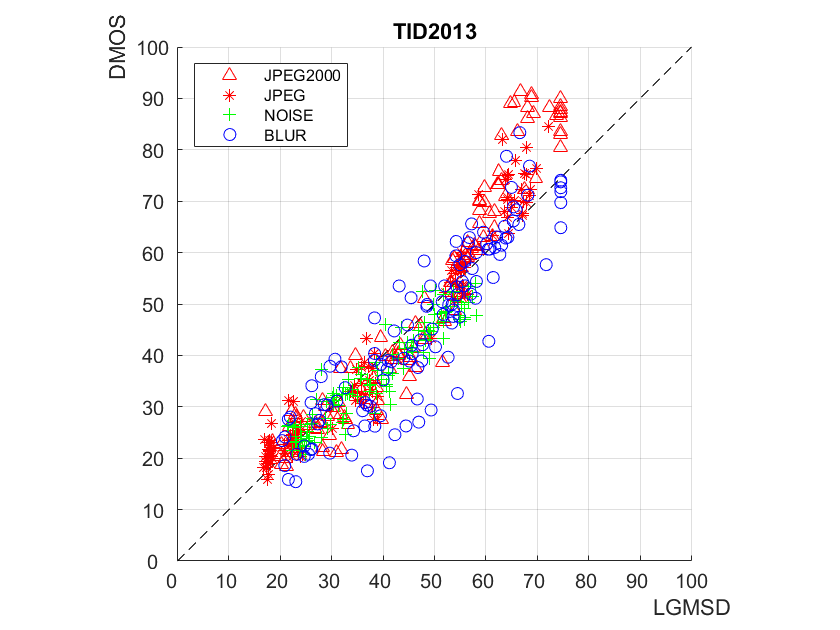}
	\includegraphics[width=1.7in]{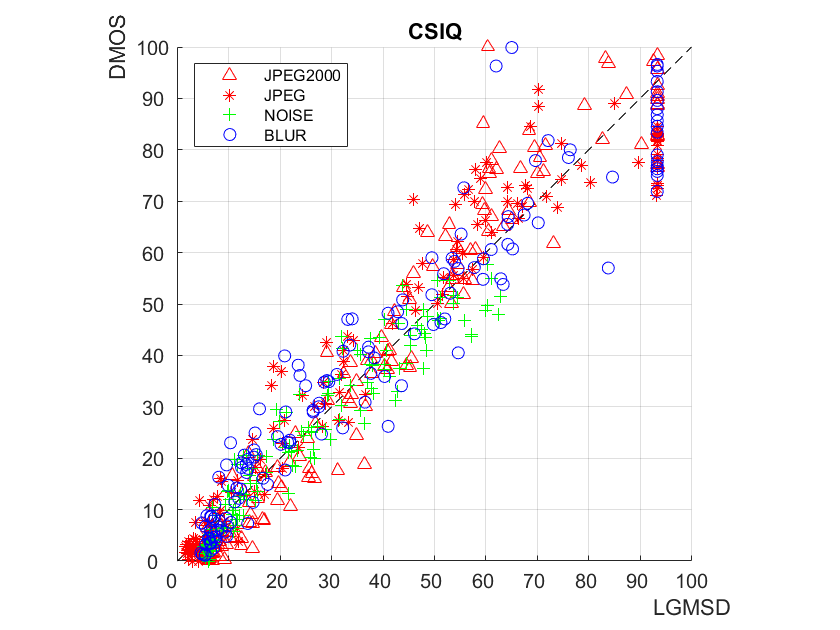}
	\includegraphics[width=1.7in]{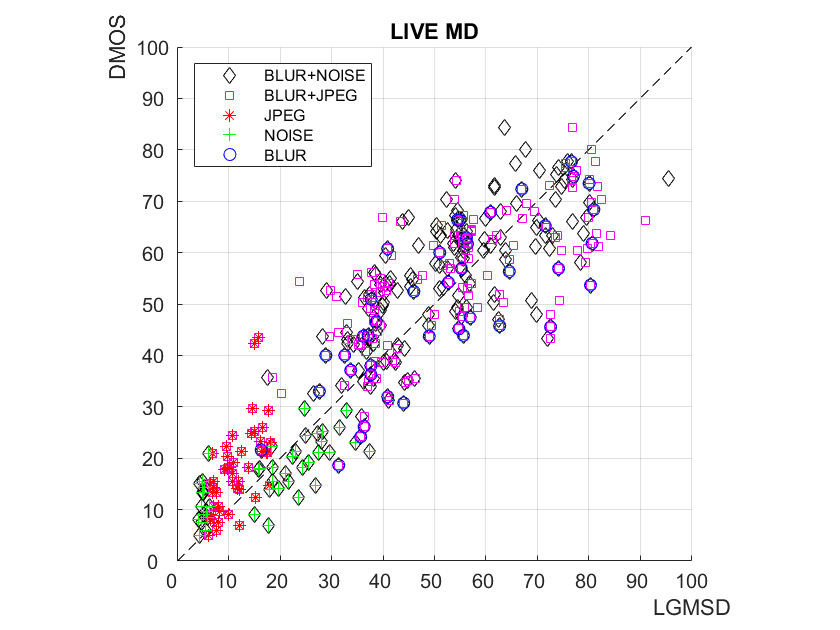}
	\includegraphics[width=1.7in]{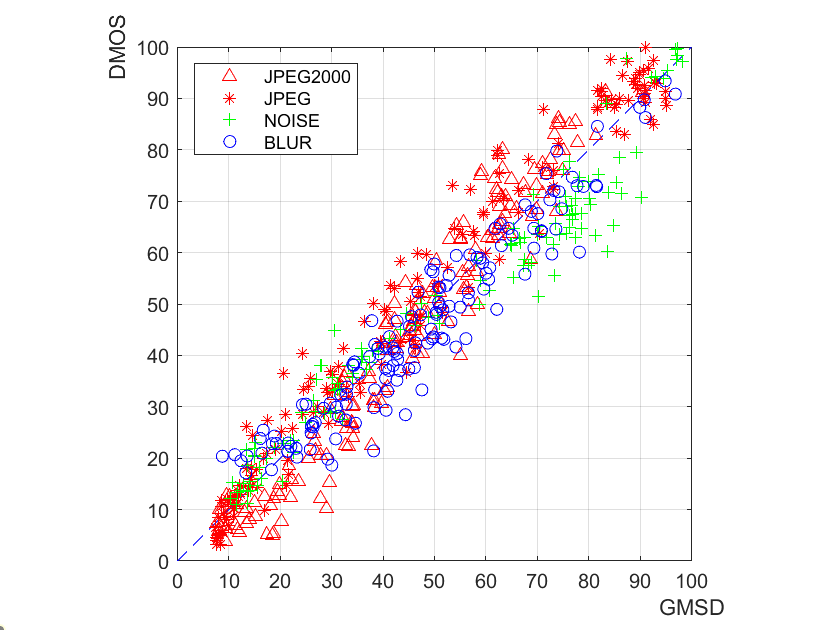}
	\includegraphics[width=1.7in]{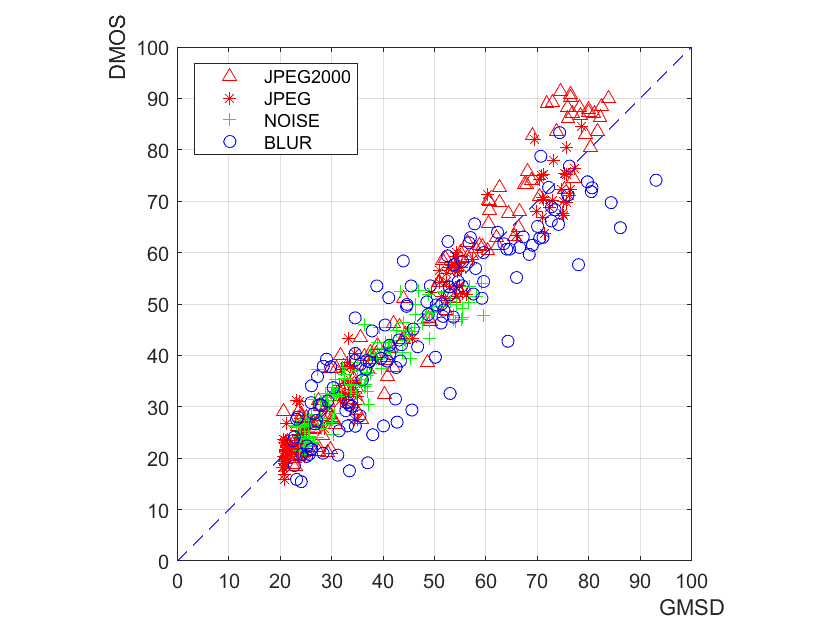}
	\includegraphics[width=1.7in]{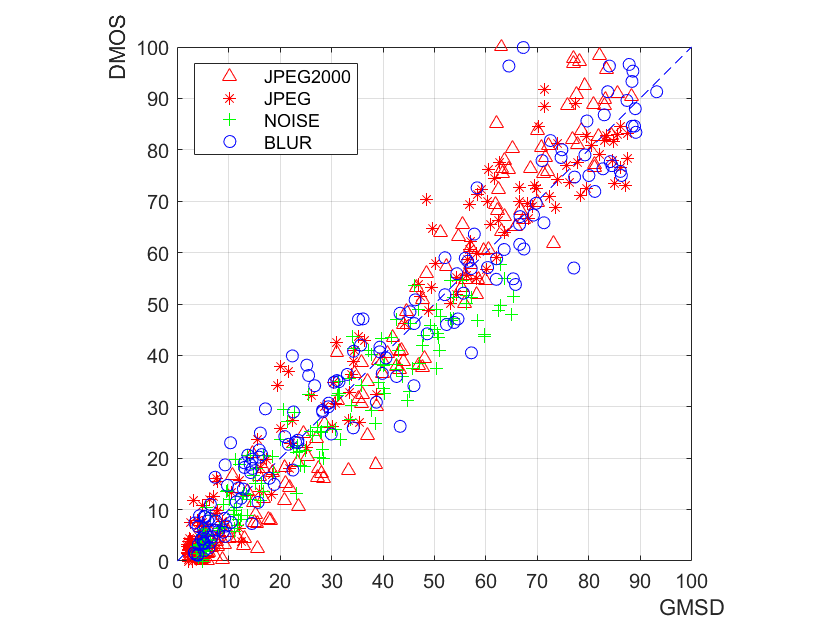}
	\includegraphics[width=1.7in]{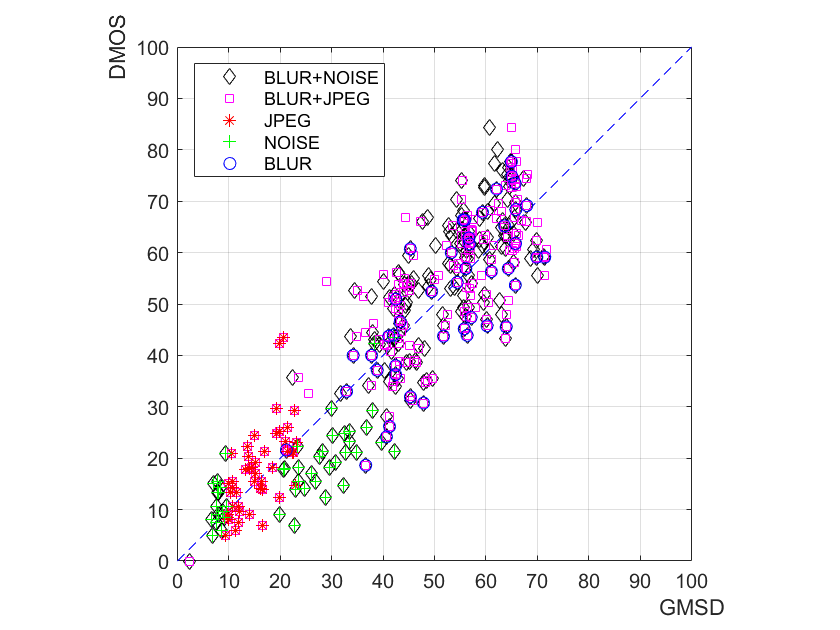}
	\caption{The DMOS scatterplots after empirical calibration for blur, noise, JPEG, JPEG2000 distortions for LIVE DBR2, TID2013 and CSIQ, and blur, noise, JPEG, blur+JPEG, blur+noise distortions for LIVE MD, versus the values predicted using respectively the theoretically linearized GMSD method (upper row) and the conventional, empirically calibrated GMSD method (lower row) for different databases.}
	\label{fig11}
\end{figure*}

Further indications come from the scatterplots of empirical DMOS values versus the estimated ones, reported in Figs.\ref{fig8},\ref{fig9},\ref{fig10},\ref{fig11}, arranged for direct visual comparison. In general, it is observed that the mutual positions of the clusters regarding the different degradations for the IQA methods tend to be preserved by the corresponding LIQA versions.

Furthermore, a slightly more visible deviation from linearity of the LIQAs for the TID2013 dataset can be noted. This could be explained by the fact that the VD in these experiments is not controlled, but left to the choice of the observers \cite{PONOMARENKO15}, whereas here it is determined by regression all over the subset of blurred image data.

\section{Implementation issues}
\noindent The computational steps for the implementation of the a-priori linearized methods are summarized as follows:

\begin{enumerate}
	\item{With reference to a typical natural image, stipulate a conventional DMOS for that image at some level of blur (\emph{anchor} value) to determine the value of $Q$, using the canonical method.}
	\item{Calculate the conversion function $\xi(\zeta;\tau)$ for the chosen conventional IQA method and for a set of VD values of interest $\tau_1, \tau_2, ..., \tau_N$. To this purpose, the images are resampled for each VD of interest.}
	\item{Calculate the equivalent blur for the image of interest starting from the metric $\zeta$ determined by the application of the chosen conventional IQA and the associated conversion function $\xi(\zeta;\tau)$ for the wanted VD.}
	\item{Apply the rating function ${\hat{d}}_{LIQA}(\zeta)$ using the already determined values of $Q$ and $\tau$.}
\end{enumerate}

Since the costly step 2 is off-line, the computational burden for the estimation of the DMOS of a given degraded image reduces substantially to that required by the calculus of the metric $\zeta$ of the selected conventional IQA method.

\section{Remarks}
\noindent In the face of the concordant results of independent experiments based on different methods, protocols and viewing devices, one might still wonder how it is possible to obtain such reasonably accurate objective quality estimates linearly correlated with subjective judgments without a-posteriori calibration. The multiplicity of factors influencing the quality judgments pushes rather toward the use of empirical analogies, as done in classical and machine learning based IQA methods. To enlighten why the present unconventional calibration-free LIQA methods do constitute a viable approach, let us further evidence that these methods are not aimed to predict the subjective quality of a specific image, but rather the expected value of a random set of images characterized by a perceptual quality equivalent to Gaussian blur, which is regarded as \emph{deterministic}. Its differences with respect to actual quality measurements are the effect of different unobserved factors. The good linear correlation of the theoretical estimates versus the empirical scores follows from the validity of the adopted general principles, namely the optimality of the HVS with respect to fine pattern localization, and the Weber-Fechner law, as discussed in Section \ref{sec:The canonical IQA method}. Just as you are certain about the time that a stone takes to fall from a given height (believing in mechanics), you should also regard such linearity as a necessary consequence of the said principles.

As outlined in Section \ref{sec:Relationship among the canonical IQA method and other IQA methods}, the most critical point is the correct determination of a specimen original image representative of the universe of natural original images, beyond the spectral fall-off property invoked in the calculus of the expected average quality.

The specimen image covers the basic role of allowing conversion between conventional quality metrics and a perceptually quality equivalent Gaussian blur. In turn, the conversion rule constitutes the bridge for converting the estimated quality among different VDs. This possibility is precluded to conventional calibration-prone methods, also because of the scarcity of empirical datasets containing experiments for different VDs \cite{LIU14}. Among others, this feature allows straightforward prediction of the subjective image quality using jointly different databases, after equalization of the VD and of $Q$.

As far as the generalization of LIQA methods to different image degradation types is concerned, it is basically inherited from the properties of the corresponding IQA methods through the conversion formulas, as deduced in Section \ref{sec:Extension of the method beyond Gaussian blur} and corroborated by the observation of the scatterplots of Section \ref{sec:Performance evaluation}.

Extending the observations made regarding the choice of the specimen image, it is also outlined that, even though the quality equivalence invoked in Section \ref{sec:Relationship among the canonical IQA method and other IQA methods} depends on empirical data through the settings of the IQA methods, these settings were established in a long series of past experiments and can be considered firmly consolidated.

Still, the methods considered in this paper cannot be strictly said “calibration-free” whenever the parameters $Q$ and $\tau$ are unknown. In practice, they have a clear operative meaning and are simply determined in applications.

\section{Conclusion}
\noindent The presented LIQA methods appear to predict with a reasonable accuracy the subjective quality loss of natural images in the FR mode for a class of technically interesting degradations. By the comparative analysis of  a degraded image against its pristine counterpart, LIQA methods provide quality loss estimates using only DMOS anchoring and VD information.

Since they conform to an abstract model of real subjects retaining their essential, shared characteristics, it is expected that they will properly reflect the statistical opinion of large populations of observers. After all, this is the substantial goal of image providers.

\bibliographystyle{IEEEtran}
\bibliography{IEEEabrv,imageprocessing}


\begin{IEEEbiography}[{\includegraphics[width=1in,height=1.25in,clip,keepaspectratio]{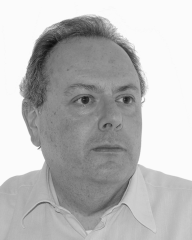}}]{Elio D. Di Claudio} received the Dr. Ing. degree in Electronics Engineering with honors from the University of Ancona, Italy, in 1986. From 1986 to 1990 he was with the Telettra S.p.A. company, Chieti, Italy, where he worked on spread-spectrum communication equipments and digital signal processing algorithms. From 1990 to 1991 he was with the ELASIS S.c.p.A. company. In 1992 he joined the INFOCOM Department of the University of Rome ``La Sapienza'' as a Researcher. From 1998 to 2007 he was Associate Professor of Circuit Theory. From 2007 he is Full Professor of Circuit Theory at the University of Rome ``La Sapienza''. Di Claudio's current interests are in the fields of array and image processing, parallel algorithms for signal processing, parallel architectures for VLSI and machine learning.
\end{IEEEbiography}

\vskip 0pt plus -1fil

\begin{IEEEbiography}[{\includegraphics[width=1in,height=1.25in,clip,keepaspectratio]{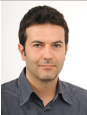}}]{Paolo Giannitrapani} received the Dr. Ing. degree in Telecommunications Engineering from the University of Rome ``Sapienza'', Italy, in 2005. From 2006 to 2007 he was with the IBM company, Rome, Italy, where he worked on new architecture on SSL/TLS layers and coding algorithms for digital signature. From 2007 to 2017 he was with the General Electric company where he worked on mechanical tool software for turbo-machinery design. From 2017 he was with Baker Hughes company as scientific lead of projects in AI and Robotics fields. His research interests include parallel algorithms for signal processing, spectral estimation, and array processing, estimation and detection theories. Main activities are in signal and image processing using Pattern Recognition and Deep Learning, linear and nonlinear optimization techniques.
\end{IEEEbiography}

\vskip 0pt plus -1fil

\begin{IEEEbiography}[{\includegraphics[width=1in,height=1.25in,clip,keepaspectratio]{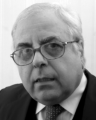}}]{Giovanni Jacovitti} is a former Full professor of Digital Signal Processing at the University of Rome ``La Sapienza''. He received the Dr. Ing. degree in Electronics Engineering in 1970 from the University of Rome. He has been also professor of Communications at the University of Cagliari and at the University of Bari, and professor of Digital Signal Processing at the University “Campus Bio-medico” of Rome. His main research activities are in the fields of Signal Processing and Communication and Estimation theories. His current interests are focused on epistemological issues about the fundamentals of information science and technologies. He promotes educational activities and conferences on these topics.
\end{IEEEbiography}

\end{document}